\documentclass[11pt]{article}

\usepackage[a4paper,margin=1in]{geometry}
\usepackage{times}
\usepackage{latexsym}
\usepackage[T1]{fontenc}
\usepackage[utf8]{inputenc}
\usepackage{microtype}
\usepackage{inconsolata}
\usepackage{graphicx}
\usepackage{booktabs}
\usepackage{amsmath}
\usepackage{amssymb}
\usepackage{multirow}
\usepackage{array}
\usepackage{tabularx}
\usepackage{threeparttable}
\usepackage{placeins}
\usepackage{xcolor}
\definecolor{ThemeBlue}{RGB}{30,90,170}
\usepackage{url}
\usepackage[numbers,sort&compress]{natbib}
\usepackage[hidelinks]{hyperref}
\hypersetup{colorlinks=true,linkcolor=ThemeBlue!85!black,citecolor=ThemeBlue!85!black,urlcolor=ThemeBlue!85!black}
\usepackage[most]{tcolorbox}
\usepackage{titlesec}
\usepackage{wrapfig}
\graphicspath{{figures/}{latex/figures/}}

\titleformat{\section}{\normalfont\Large\bfseries}{\thesection.}{0.5em}{}
\titleformat{\subsection}{\normalfont\large\bfseries}{\thesubsection}{0.5em}{}

\makeatletter
\renewcommand{\maketitle}{%
  \begingroup
    \begin{flushleft}
      {\huge\bfseries\@title\par}
      \vspace{1.0em}
      {\normalsize\@author\par}
    \end{flushleft}
  \endgroup
  \vspace{1.0em}
}
\makeatother

\renewenvironment{abstract}{%
  \begin{tcolorbox}[
    enhanced,
    colback=ThemeBlue!7,
    colframe=ThemeBlue!7,
    arc=2pt,
    boxrule=0pt,
    left=14pt, right=14pt, top=10pt, bottom=10pt,
  ]
  \small\noindent\ignorespaces
}{%
  \end{tcolorbox}
  \vspace{0.6em}
}

\usepackage{enumitem}

\usepackage{colortbl}
\usepackage{listings}
\lstdefinestyle{tronpy}{
  language=Python,
  basicstyle=\scriptsize\ttfamily,
  keywordstyle=\color{blue!60!black}\bfseries,
  stringstyle=\color{red!55!black},
  commentstyle=\color{green!40!black}\itshape,
  showstringspaces=false,
  columns=fullflexible,
  keepspaces=true,
  breaklines=true,
  breakatwhitespace=false,
  frame=single,
  rulecolor=\color{black!30},
  framesep=3pt,
  xleftmargin=10pt,
  framexleftmargin=4pt,
  numbers=left,
  numberstyle=\tiny\color{black!40},
  numbersep=4pt,
  aboveskip=3pt,
  belowskip=3pt,
  abovecaptionskip=2pt,
  belowcaptionskip=2pt,
}
\definecolor{DeltaBlue}{RGB}{35,92,160}
\newcommand{\bench}[1]{{\fontsize{6.7}{7.2}\selectfont\textbf{#1}}}
\newcommand{\dnum}[1]{\textcolor{DeltaBlue}{#1}}

\newcommand{\tron}{\textsc{TRON}}
\newcommand{\method}{\textsc{TRON-DAPO}}

\DeclareRobustCommand{\tronlogo}{%
  \raisebox{-0.12em}{\includegraphics[height=1.05em]{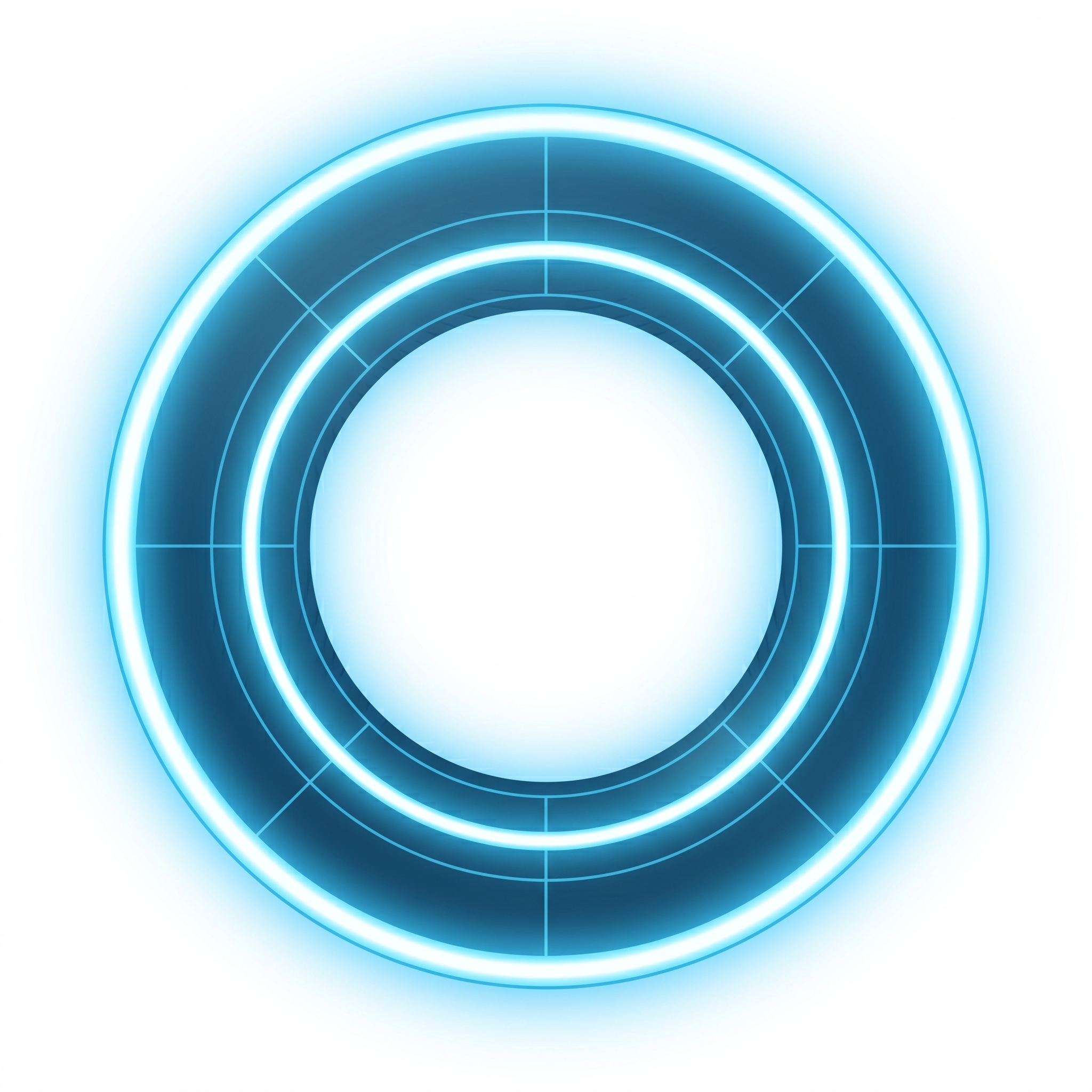}}%
}

\title{\textcolor{ThemeBlue}{\tronlogo\hspace{0.35em}\tron:} Targeted Rule-Verifiable Online Environments for Visual Reasoning RL}

\author{%
  Tianze Yang\textsuperscript{*}\quad Yucheng Shi\textsuperscript{*}\quad Ruitong Sun\quad Jingyuan Huang\quad Ninghao Liu\quad Jin Sun\\[2pt]
  University of Georgia\\[4pt]
  {\normalsize\textbf{Project page:} \textcolor{ThemeBlue}{\url{https://tron-rl.github.io/}}}\\[3pt]
  {\footnotesize\textsuperscript{*}Equal contribution.}
}

\begin{document}
\maketitle

\begin{abstract}
Reinforcement learning (RL) for visual reasoning needs scalable,
verifiable, and controllable training signals.  Existing visual RL
post-training trains on \emph{static} curated datasets, with fixed
image-question-answer samples bounded by their collection
budget. In this work, we introduce \tron{} (\emph{Targeted, Rule-verifiable Online
eNvironments}), an online environment substrate: a training
rollout is generated on demand by a controllable generator--verifier
program that samples a fresh latent visual state, renders an image,
asks a question, and exactly verifies the answer. A single run can therefore
draw an unbounded stream of fresh instances at the difficulty level required by the current curriculum.  The current TRON suite contains 520
environments organized into five ability buckets (spatial,
mathematical, diagram, pattern/logic, and counting); the same
substrate supports both a single \emph{full} model trained on all
buckets and per-bucket \emph{ability-specialist} models, with no
additional data collection.  We also introduce a substrate
analysis covering generation reliability, instance and level
diversity, cross-environment near-duplicates, and base-model pass
rate by difficulty level.  RL post-training with \method{}
consistently improves performance on ten external multimodal reasoning benchmarks across Qwen3-VL-4B, Qwen2.5-VL-7B, and MiMo-VL-7B.
\end{abstract}

\section{Introduction}
\label{sec:intro}

Recent multimodal language models increasingly rely on reward-based
post-training. In domains such as reasoning, mathematics and code, RL can often exploit
exact supervision: a numeric answer can be deterministically checked~\citep{shao2024deepseekmath,guo2025deepseekr1,lambert2024tulu3},
a program can be executed against hidden tests~\citep{hendrycks2021apps,
li2022alphacode}, and a proof can be machine-verified by a kernel
\citep{yang2023leandojo,zheng2022minif2f}.
However, visual reasoning is substantially harder. A model may need to count
occluded objects, infer spatial relations, trace diagrams, interpret charts, or solve visually grounded puzzles.  These tasks are easy to
package as evaluation examples \citep{qiao2024wemath,zhang2024mathverse,
wang2024spatialeval,xiao2024logicvista}, but difficult to turn into scalable RL training signals, because each instance needs a visual scene, an unambiguous question, a calibrated difficulty, and a reliable verifier.

Existing visual reasoning training pipelines rely on
image-question-answer datasets, collected either through human
annotation \citep{qiao2024wemath,zhang2024mathverse,masry2025chartqapro}
or synthetic instruction generation
\citep{zhang2024mavis,shi2024mathllava,gao2023gllava,
han2023chartllama,masry2024chartgemma}.  This abstraction works for
evaluation but is a weak fit for RL training. 
First, static datasets are finite and costly to annotate, so the dataset size is bounded by the curation budget rather than by what the model could productively consume.
Second, it provides little control over the specific skill being practiced or the difficulty presented to the model at different stages of training. 
Third, as newer VLMs absorb many popular reasoning datasets during pretraining and supervised fine-tuning, these datasets become less useful as RL training signals because the model has often already seen substantial portions of them.

We therefore take a different approach: visual reasoning RL should train on a \textbf{diverse suite of procedural environments} rather than on a fixed collection of static VQA examples.  
We propose \tron{} (Targeted, Rule-verifiable Online
eNvironments), an online visual reasoning substrate in which each environment generates fresh training instances together with exact rewards.
A \tron{} environment owns both a generator and a verifier. The generator samples a latent visual state, renders an image, and constructs a question; the verifier computes and checks the correct answer from the same state.
During RL training, the model observes only the image and the question, while the reward is provided by the deterministic verifier.
This formulation allows the training process to target specific reasoning mechanisms directly. For operations such as chart aggregation, cube rotation, occluded counting, 
visual analogy, or graph search, 
we instantiate task environments centered on those operations.

This procedural formulation turns data generation into a controllable mechanism. 
The \textbf{diversity} of generated data is built on three levels.
First, different environments target different reasoning mechanisms. Second, each environment generates distinct visual instances by varying layouts, objects, and distractors. Third, each environment has a difficulty ladder that produces progressively harder versions of the same operation.
As the model improves, the training source does not become exhausted. The environments continue to generate fresh rollouts at an appropriate level of challenge.


However, generating large numbers of samples does not automatically produce useful training signal. For example, a generator may vary superficial metadata while leaving the underlying task unchanged, different environments may collapse to the same effective template, or a verifier may incorrectly accept invalid answers. 
\tron{} therefore couples controllable generation with a substrate audit that checks rendering and verifier correctness, measures diversity across instances and difficulty levels, detects near-duplicate environments, and verifies that higher difficulty levels correspond to genuinely harder tasks for the base model.

This paper makes the following contributions:
\begin{enumerate}[noitemsep, topsep=0pt, leftmargin=*]
  \item We introduce \tron{}, an \emph{online} environment substrate
  for visual reasoning RL: 520 generator-verifier programs that
  produce fresh image-question rollouts at training time, with no
  cap on instances per run.
  \item We organize the substrate into five ability buckets, and use it
  to train both a single \emph{full} model and
  per-bucket \emph{ability-specialist} models without additional data collection.
  Our analysis reveals new insights on ability transfer.
  \item We evaluate generation quality, diversity, and difficulty calibration, showing that RL training with \tron{} consistently improves different open VLM families, including Qwen3-VL-4B-Instruct~\citep{qwen2025qwen3vl},
  Qwen2.5-VL-7B-Instruct~\citep{bai2025qwen25vl}, and
  MiMo-VL-7B-SFT~\citep{yue2025mimo}, across ten reasoning benchmarks.
\end{enumerate}

\begin{figure*}[t]
\centering
\includegraphics[width=0.99\textwidth]{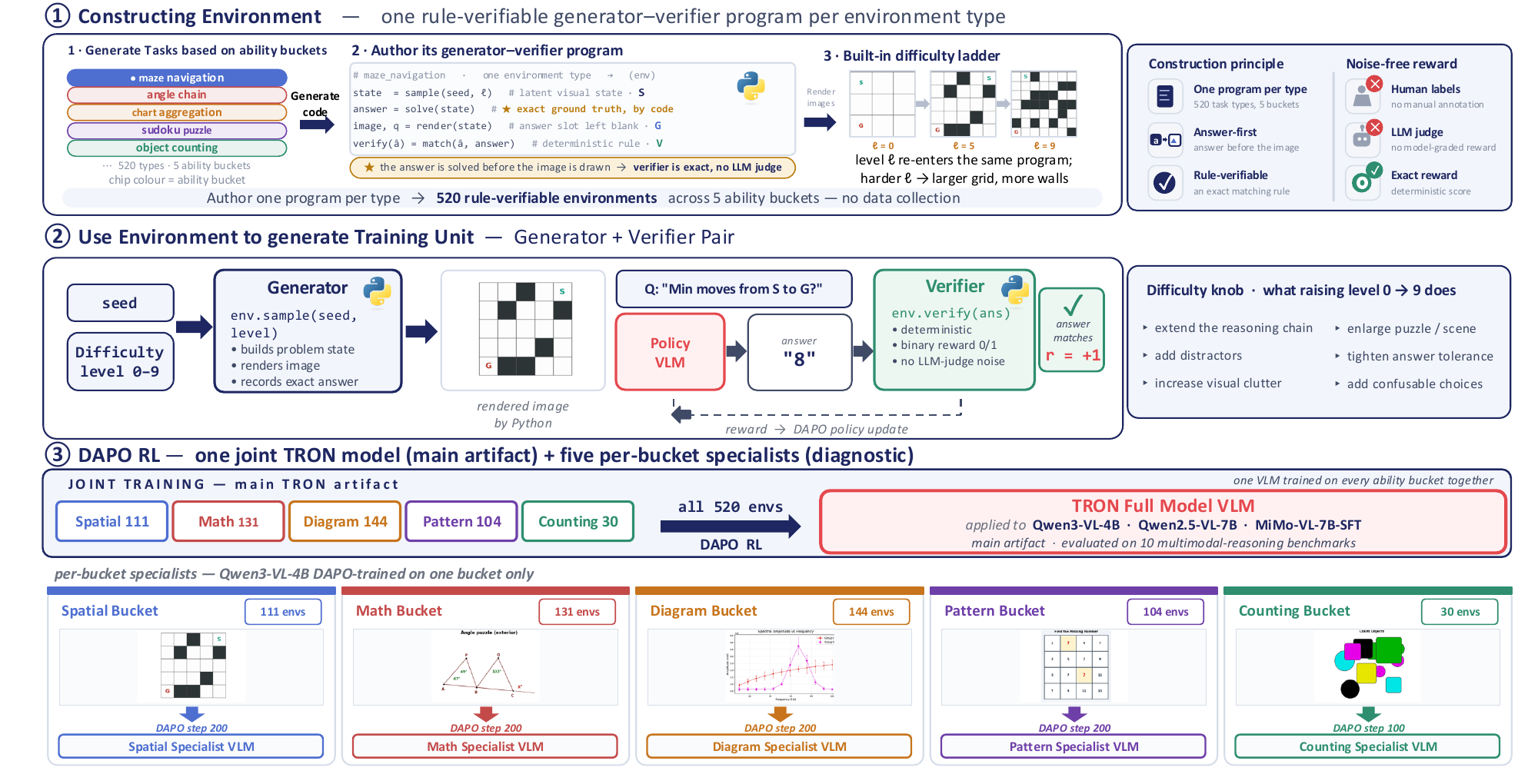}
\caption{\textbf{\tron{}: diverse, ability-targeted, auditable environments for visual reasoning RL.}  \tron{} organizes 520 rule-verifiable generators into ability buckets covering spatial, mathematical, diagram, pattern, and counting skills.  Each environment produces fresh difficulty-controlled image--question rollouts with a deterministic verifier; a substrate analysis (Section~\ref{sec:audit}) checks generation quality, instance and level diversity, cross-environment near-duplicates, and base-model pass rate by difficulty level before mixed or ability-specialist RL training.}
\label{fig:tron_overview}
\end{figure*}

\section{Related Work}
\label{sec:related}
\noindent\textbf{RLVR and visual RL post-training.} Reinforcement learning with verifiable rewards (RLVR) has become a central recipe for improving language-model reasoning \citep{lambert2024tulu3,shao2024deepseekmath,guo2025deepseekr1}, with variants such as GRPO and DAPO refining the optimization \citep{shao2024deepseekmath,yu2025dapo}. A parallel line extends RLVR to vision-language models on mathematical, spatial, grounding, counting, and multimodal reasoning tasks \citep{liu2025visualrft,shen2025vlmr1,meng2025mmeureka,huang2025visionr1,peng2025lmmr1,chen2025r1v,yang2025r1onevision,wang2025vlrethinker,leng2025mmr1,tan2025reasonrft}. These methods share our use of rule-based rewards but train on \emph{static} curated data; \tron{} replaces this static substrate with online generator--verifier environments that produce fresh instances and advance a per-environment curriculum on demand.

\noindent\textbf{Procedural generator--verifier environments.} Procedural generation has long been used to improve RL generalization \citep{cobbe2020procgen}. Text-only systems such as Reasoning Gym, SynLogic, Enigmata, and the curriculum-driven RLVE framework \citep{stojanovski2025reasoninggym,liu2025synlogic,chen2025enigmata,li2024numinamath,zeng2025rlve} show that procedurally generated rule-verifiable tasks provide scalable RL signal beyond fixed math corpora, and the same principle underlies code and formal-math verifiers \citep{hendrycks2021apps,li2022alphacode,yang2023leandojo,zheng2022minif2f}. \citet{meng2026gym} extends procedural environments to the visual setting via agentic interaction over a small set of tasks; \tron{} instead provides a much more diverse, image visual reasoning suite of 520 environments, each rendering images grounded in latent visual states.

\noindent\textbf{Synthetic visual reasoning and capability decomposition.} Synthetic visual reasoning benchmarks demonstrate the value of controlled visual states and explicit reasoning programs: CLEVR introduces functional-program supervision for counting, comparison, existence, and spatial relations \citep{johnson2017clevr}; RAVEN and procedurally generated matrices study visual analogy and abstract pattern completion \citep{zhang2019raven,barrett2018pgm}; Bongard and ARC-style tasks emphasize rule induction and skill acquisition \citep{nie2020bongardlogo,chollet2019arc}. Contemporary multimodal benchmarks further decompose VLM performance into mathematical, spatial, chart, diagram, logic, and puzzle-oriented capabilities \citep{qiao2024wemath,zhang2024mathverse,zou2024dynamath,wang2024spatialeval,xiao2024logicvista,yuan2025mmereasoning,wang2024charxiv,masry2025chartqapro,chia2024puzzlevqa}. \tron{} draws on this decomposition as an authoring guide but builds a reusable RL training suite rather than another evaluation taxonomy: 520 environments generate fresh image--question rollouts with deterministic verifiers, local difficulty ladders, and the substrate analysis of Section~\ref{sec:audit}.

\section{\tron{} Environments}
\label{sec:envs}

This section presents the \tron{} framework:
Section~\ref{sec:env_definition} defines an environment as a generator--verifier pair that produces noise-free RL signals, and Section~\ref{sec:suite} organizes 520 environments into five reasoning categories with per-environment difficulty ladders.

\subsection{Environment Definition}
\label{sec:env_definition}

A \tron{} environment $e$ is a tuple
$(\mathcal{S},\,\mathcal{L},\,G,\,V)$, where $\mathcal{S}$ is an
\emph{underlying task state} used to generate problem instances 
(e.g.\ cube
configurations, chart data tables, or puzzle solutions), and $\mathcal{L}=\{0,1,\ldots,9\}$ is a set of \emph{difficulty levels} with higher $\ell$ producing
harder instances of the same underlying mechanism.  
The
\emph{generator}
$
G:\mathcal{S}\!\times\!\mathcal{L}\to
\mathcal{I}\!\times\!\mathcal{Q}\!\times\!\mathcal{A}
$
deterministically maps a sampled state $s\in\mathcal{S}$ and a level
$\ell\in\mathcal{L}$ to a rendered image $I\in\mathcal{I}$, a
natural-language question $q\in\mathcal{Q}$, and the ground-truth
answer $a\in\mathcal{A}$.  
The \emph{verifier}
$V:\mathcal{A}\!\times\!\mathcal{A}\to\mathbb{R}$ takes the
ground-truth $a$ and a model prediction $\tilde a\in\mathcal{A}$
and returns a scalar score that reflects correctness under the
environment's matching rule (exact comparison, set or sequence
equality, or a puzzle-specific solver) together with any optional
format signal.  At training time the policy observes only $(I,q)$
and receives reward $V(a,\tilde a)$.

Three properties distinguish our formulation from the static
image--question--answer triples used by prior visual RL datasets:
1) Dataset size is bounded only by training compute rather than by a
curation budget; 2) No fixed item set exists for a successor VLM to
absorb during pretraining or SFT, and 3) The difficulty parameter
$\ell$ is a curriculum-controlled input set per \textit{sample} rather than a set-and-forget artifact.

\subsection{Environment Construction}
\label{sec:suite}

Each of the 520 \tron{} environments is a Python program targeting
one class of reasoning mechanism. Our choice of axes follows from a
survey of contemporary visual-reasoning benchmarks and
datasets~\citep{lu2024mathvista,wang2024spatialeval,wang2024charxiv,xiao2024logicvista,chia2024puzzlevqa,yuan2025mmereasoning},
from which we distil the core abilities that strong large
vision-language models are expected to handle reliably; the suite
covers five such high-level ability axes (spatial, math, diagram,
pattern/logic, counting; see Table~\ref{tab:buckets} and
Appendix~\ref{app:capability_map}).

Producing one training instance from environment $\text{env}$ at
level $\ell$ takes five steps:
(i)~the $(\text{env}, \ell)$ pair determines a specific problem
type --- for the simplest angle-chase level, the type is ``two
interior angles of a triangle are given, find the third'';
(ii)~sample the type's free variables from a random seed (here,
the two known angles, e.g.\ $a\!=\!55^\circ$ and $b\!=\!70^\circ$);
(iii)~apply a pre-defined formula or solver to compute the answer
(here, $c\!=\!180^\circ\!-\!a\!-\!b\!=\!55^\circ$);
(iv)~render the problem into an image with the answer slot left
blank (here, a triangle with $a$ and $b$ labelled and the third
corner marked ``$x\!=\!?$'');
(v)~sample the question wording from a small pool of paraphrases.
Because the answer is fixed before the image is drawn, the verifier
always holds the unique correct value and never needs to parse the
rendered image, so the RL reward is exact.

The level $\ell$ switches the mechanism inside the same
environment: in angle-chase, $\ell$ scales the number of geometric
deduction steps (one-step triangle sum at $\ell\!=\!0$; four-step
composite chain over parallels at $\ell\!=\!9$); in
chart-aggregation, $\ell$ scales the number of series, the number
of time points, and whether series are stacked. 

Diversity therefore enters at three places: within an environment
the seed randomises the latent state (values, layout, palette,
label names) and the question stem; across levels $\ell$ shifts
the mechanism; across environments the mechanism class itself
changes.  Section~\ref{sec:audit} audits all $3$ dimensions,
and Appendix~\ref{app:qual_examples} shows sampled rendered
examples.

\begin{table}[t]
\centering
\small
\setlength{\tabcolsep}{4pt}
\renewcommand{\arraystretch}{1.08}
\begin{tabularx}{\columnwidth}{@{}l r X@{}}
\toprule
\textbf{Bucket} & \textbf{\#} & \textbf{Core mechanisms} \\
\midrule
Spatial & 111 & 3D rotation, cube nets and folding, navigation and pathfinding, perspective shifts, mechanical layout \\
Math & 131 & Geometric theorems (angles, circles, polygons), analytic geometry, algebra, probability over visual quantities \\
Diagram & 144 & Chart aggregation (bar/line/scatter), tables, graph algorithms (trees, Eulerian/Hamiltonian, max-flow), flowcharts, scientific figures \\
Pattern/Logic & 104 & Constraint puzzles (sudoku, binairo, calcudoku), visual analogies, syllogistic deduction, sequence completion, state-space planning (Hanoi, Sokoban) \\
Counting & 30 & Visual enumeration of objects, cells, and regions; path counting; measurement and feature estimation \\
\bottomrule
\end{tabularx}
\caption{Compact overview of the 520-environment \tron{} suite. Each bucket groups rule-verifiable generator--verifier programs around reusable visual reasoning mechanisms; Appendix~\ref{app:capability_map} provides the fine-grained environment map.}
\label{tab:buckets}
\end{table}


\section{RL Post-Training}
\label{sec:training}

\paragraph{Data generation.}
Training data is produced online from the \tron{} substrate rather
than from a fixed corpus.  At every sampling step the trainer picks a
tuple $(\text{env}, \ell, \text{seed})$ and invokes the environment's
recipe of Section~\ref{sec:suite} to obtain a fresh $(I, q, a)$;
because the seed is fresh each call, no two training steps see the
same instance.  At training time the image $I$ is additionally
perturbed to improve robustness: every sample receives a small
white-pad size jitter ($0$--$40$ pixels per side), and with
probability $0.30$ one perturbation is drawn from \{rotation
$\pm 3$--$8^\circ$, low-quality JPEG, brightness shift, Gaussian blur,
additive Gaussian noise\}.  Each environment carries its own
difficulty level $\ell$, which, following~\citet{zeng2025rlve}, we
couple to the rollout stream rather than to offline epochs: we track
recent verifier accuracy at the
current $\ell$ and promote the environment to $\ell\!+\!1$ once that
accuracy crosses a threshold over a target number of graded
trajectories, while a sliding window over recent levels retains
lower-level skills.  The same 520-environment substrate supports both
a single \emph{full} model trained across all buckets and per-bucket
\emph{ability-specialist} models (Section~\ref{sec:ability_specialists})
via a sampler configuration switch, with no extra data.

\paragraph{Training recipe.}
We optimize with a DAPO-style objective~\citep{yu2025dapo} and
prompt-grouped advantages in the spirit of
GRPO~\citep{shao2024deepseekmath}: the rollout engine draws $n=8$
responses for each prompt, the environment's deterministic verifier
scores each response $\tilde a$, and the RL reward is $V(a, \tilde
a)$.  The reward is computed without an LLM judge; each verifier
checks the answer and the requested wrapper format, with lightweight
numeric or symbolic normalization handled inside the verifier when
appropriate.  We use DAPO clip-higher (low/high clip ratio $0.2/0.28$)
for exploration on negative samples and group filtering to drop
prompt groups that are all-correct or all-wrong from the policy
update; group filtering happens \emph{after} verifier scoring, so
uninformative rollouts still contribute their signal to the
per-environment curriculum accumulator.  KL regularization uses a
low-variance estimator with coefficient $0.005$ and is not added to
the reward, and the entropy coefficient is set to $0$.  All runs are
trained on a single node of $4\times$ H100 80\,GB GPUs with vLLM
tensor parallelism $4$.  Full hyperparameters (batch sizes, learning
rate, and curriculum thresholds) for the full and ability-specialist
runs are listed in Appendices~\ref{app:full_training_details}
and~\ref{app:ability_training_details}.

\section{Experiments}
\label{sec:experiments}
We first audit the environments in~\tron{} (Sec.~\ref{sec:audit}) on quality, difficulty, and diversity. Then we show the main results of training state-of-the-art visual reasoning models on~\tron{} and evaluate on external benchmarks (Sec.~\ref{sec:benchmarks} and~\ref{sec:main_results}). Finally, we present ability-specialist results (Sec.~\ref{sec:ability_specialists}), utilizing the ability partition in~\tron. 

\subsection{Environment Analysis}
\label{sec:audit}

\begin{wrapfigure}{r}{0.52\textwidth}
\centering
\vspace{-\intextsep}
\includegraphics[width=0.5\textwidth]{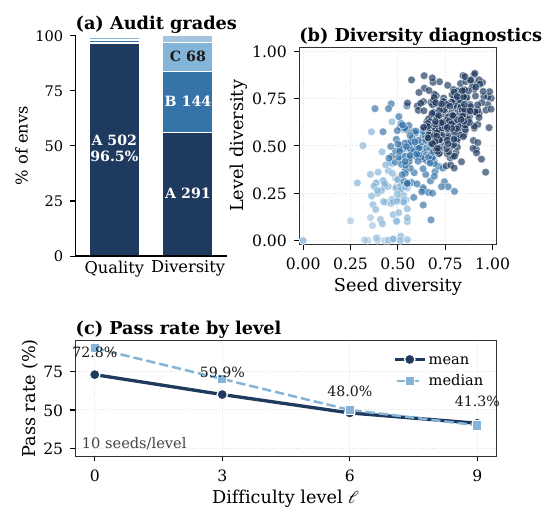}
\caption{Model-free audit of the 520 training environments (4 levels
$\times$ 4 seeds = 8{,}320 probes, 99.1\% success).  (a) Quality and
diversity grade distributions.  (b) Per-environment seed vs.\ level
diversity, colored by overall diversity grade.  (c) Qwen3-VL-4B base
pass rate on the same audited levels.}
\label{fig:audit_summary}
\end{wrapfigure}

We empirically validate the 520-environment substrate before any RL
training.  Three silent failure modes would compromise it:
(i)~the generator--verifier pair emits a malformed probe (blank
render, missing fields, dropped sample) or admits a wrong answer,
corrupting the per-sample reward;
(ii)~raising the difficulty level does not actually make the problem
harder for the model, leaving the curriculum without a real axis;
(iii)~seeds within an environment produce nearly identical samples,
or two distinctly-named environments collapse to nearly the same
image--question distribution.
We address these with the measurements summarized in
Figure~\ref{fig:audit_summary}.  The audit samples four levels
$\ell\in\{0,3,6,9\}$ and four seeds per level ($8{,}320$ probes) and
grades each environment A/B/C/D.  For mode~(i), a quality score
$Q(e)$ checks generation success, render validity, field presence,
and verifier sanity.  For mode~(ii), which is a semantic property
the static audit cannot test directly, we additionally run the
Qwen3-VL-4B on the same four levels with ten seeds per level
and read off its pass-rate curve (panel c).  For mode~(iii), a
diversity score $D(e)$ combines intra-environment spread
components (across seeds and across levels) with a
cross-environment near-duplicate predicate.  We describe $Q(e)$ and $D(e)$ below.

\paragraph{Quality score.}
Let $\mathcal{P}_e$ be the requested probes and $\mathcal{O}_e
\subseteq \mathcal{P}_e$ the successfully generated ones.  $Q(e)$
is the worst case across four pass rates: generation success
$r_\mathrm{gen}=|\mathcal{O}_e|/|\mathcal{P}_e|$, valid rendering
$r_\mathrm{img}$ (fraction of $\mathcal{O}_e$ with non-trivial image
size and foreground content), non-empty question and answer fields
$r_\mathrm{qa}$ (fraction of $\mathcal{O}_e$ with both fields
populated), and verifier sanity $r_\mathrm{ver}$ (fraction of
$\mathcal{O}_e$ where the wrapped correct answer is accepted and a
fixed wrong payload is rejected):
\begin{equation}
\label{eq:quality_score}
Q(e) = \min\{r_\mathrm{gen},\,r_\mathrm{img},\,r_\mathrm{qa},\,
r_\mathrm{ver}\}.
\end{equation}
Thresholds $\{0.98,0.90,0.75\}$ assign grades A/B/C, with D below.
Generation succeeds for $99.07\%$ of probes; $502/520$ environments
($96.5\%$) receive grade~A and the 18 below-A environments were
re-authored.
Gate predicates are in
Appendix~\ref{app:quality_gate_impl}.

\paragraph{Difficulty.}
Figure~\ref{fig:audit_summary}(c) confirms that the difficulty axis
is valid for the base model: running Qwen3-VL-4B-Instruct on the same
four levels with ten seeds per level, the mean pass rate across the
suite falls from $72.8\%$ at $\ell=0$ to $59.9\%$
($\ell=3$), $48.0\%$ ($\ell=6$), and $41.3\%$ ($\ell=9$), with the
median following the same shape.  The $\approx 31$\,pp drop shows
that higher levels are not merely nominally relabeled but actually
present harder problems, so the curriculum has a real
axis to advance along.

\begin{table*}[t]
\centering
\footnotesize
\setlength{\tabcolsep}{1.7pt}
\renewcommand{\arraystretch}{1.06}
\begin{threeparttable}
\begin{tabular*}{\textwidth}{@{\extracolsep{\fill}}llrrrrrrrrrrrr@{}}
\toprule
\textbf{Model} & \textbf{Run} & \multicolumn{4}{c}{\textbf{Mathematical reasoning}} & \multicolumn{4}{c}{\textbf{Spatial and logical reasoning}} & \multicolumn{3}{c}{\textbf{Charts, figures, puzzles}} & \textbf{Mean} \\
\cmidrule(lr){3-6} \cmidrule(lr){7-10} \cmidrule(lr){11-13} \cmidrule(l){14-14}
& & \bench{WeMath-S} & \bench{WeMath-L} & \bench{MathV} & \bench{Dyna} & \bench{MME-R} & \bench{Spat.} & \bench{Logic} & \bench{HELIX} & \bench{CharXiv} & \bench{ChartQA} & \bench{Puzzle} & \\
\midrule
\multirow{3}{*}{Qwen3 4B} & Base & 52.86 & 70.95 & 43.15 & 66.11 & 42.85 & 77.52 & 57.05 & 21.67 & 39.80 & 34.41 & 72.35 & 52.61 \\
& +\tron & 58.29 & 74.86 & 44.54 & 67.49 & 45.88 & 80.47 & 59.73 & 25.56 & 42.40 & 35.04 & 73.25 & 55.23 \\
\rowcolor{black!4}
& \textcolor{DeltaBlue}{$\Delta$} & \dnum{+5.43} & \dnum{+3.91} & \dnum{+1.39} & \dnum{+1.38} & \dnum{+3.03} & \dnum{+2.95} & \dnum{+2.68} & \dnum{+3.89} & \dnum{+2.60} & \dnum{+0.63} & \dnum{+0.90} & \textbf{\dnum{+2.62}} \\
\midrule
\multirow{3}{*}{Qwen2.5 7B} & Base & 36.10 & 55.71 & 43.55 & 53.55 & 26.60 & 57.67 & 46.98 & 4.44 & 37.40 & 40.50 & 46.80 & 40.85 \\
& +\tron & 39.24 & 58.48 & 46.50 & 55.35 & 28.62 & 62.03 & 47.43 & 5.16 & 38.90 & 42.64 & 52.55 & 43.35 \\
\rowcolor{black!4}
& \textcolor{DeltaBlue}{$\Delta$} & \dnum{+3.14} & \dnum{+2.77} & \dnum{+2.95} & \dnum{+1.80} & \dnum{+2.02} & \dnum{+4.36} & \dnum{+0.45} & \dnum{+0.72} & \dnum{+1.50} & \dnum{+2.14} & \dnum{+5.75} & \textbf{\dnum{+2.50}} \\
\midrule
\multirow{3}{*}{MiMo 7B} & Base & 62.10 & 80.57 & 70.89 & 74.37 & 45.29 & 78.77 & 63.31 & 26.19 & 58.70 & 59.65\tnote{1} & 77.20 & 63.37 \\
& +\tron & 68.86 & 82.19 & 73.65 & 76.23 & 46.46 & 86.58 & 66.89 & 30.32 & 62.60 & 60.34\tnote{1} & 77.35 & 66.50 \\
\rowcolor{black!4}
& \textcolor{DeltaBlue}{$\Delta$} & \dnum{+6.76} & \dnum{+1.62} & \dnum{+2.76} & \dnum{+1.86} & \dnum{+1.17} & \dnum{+7.81} & \dnum{+3.58} & \dnum{+4.13} & \dnum{+3.90} & \dnum{+0.69} & \dnum{+0.15} & \textbf{\dnum{+3.13}} \\
\bottomrule
\end{tabular*}
\begin{tablenotes}[flushleft]
\scriptsize
\item Model abbreviations: Qwen3 4B = Qwen3-VL-4B-Instruct; Qwen2.5 7B = Qwen2.5-VL-7B-Instruct; MiMo 7B = MiMo-VL-7B-SFT.
\item[1] Format-normalized ChartQA Pro rejudge for MiMo; raw MiMo outputs do not match the benchmark parser.
\end{tablenotes}
\end{threeparttable}
\caption{
Main result: training on~\tron{} consistently improves three SOTA VLM models on all external benchmarks.
}
\label{tab:main_results}
\end{table*}

\paragraph{Diversity score.}
$D(e) = w_s D_s(e) + w_l D_l(e) + w_x D_x(e)$ scores diversity along
three axes.
\emph{Seed} diversity $D_s$ aggregates same-level perceptual-hash
(pHash) spread, question-template fraction, and answer entropy, and
checks whether seeds at one level yield visibly different problems;
low values flag near-identical seeds.
\emph{Level} diversity $D_l$ aggregates cross-level pHash distance,
template-Jaccard distance, and foreground-complexity shift over
adjacent audited pairs, and checks whether higher levels present
materially different inputs; low values flag knobs that change only
hidden metadata.
\emph{Cross-environment} diversity $D_x \in \{0,1\}$ is a
near-duplicate indicator combining pHash, thumbnail, and
prompt-template similarity, and flags environments that collapse to
a near-duplicate of another.
Appendix~\ref{app:diversity_details} gives the explicit formulas,
combination weights $w_s, w_l, w_x$, and A--D grade thresholds.
In the sweep, $435/520$ environments ($83.7\%$) receive grade A or
B; the suite-wide medians of $D_s$ and $D_l$ are $0.690$ and $0.541$.

\subsection{Benchmarks}
\label{sec:benchmarks}

We evaluate on external multimodal reasoning benchmarks covering
mathematics, spatial reasoning, chart understanding, scientific
figures, visual puzzles, and logical reasoning.  The evaluation suite
includes WeMath~\citep{qiao2024wemath},
MM-HELIX~\citep{zhao2025mmhelix},
MME-Reasoning~\citep{yuan2025mmereasoning},
SpatialEval~\citep{wang2024spatialeval},
LogicVista~\citep{xiao2024logicvista},
CharXiv~\citep{wang2024charxiv},
MathVerse-Mini~\citep{zhang2024mathverse},
DynaMath~\citep{zou2024dynamath},
PuzzleVQA~\citep{chia2024puzzlevqa}, and
ChartQA Pro~\citep{masry2025chartqapro}.  All scores are percentages
from VLMEvalKit~\citep{duan2024vlmevalkit} or official accuracy-like
outputs.

\subsection{Main Results}
\label{sec:main_results}

Table~\ref{tab:main_results} shows that \tron{} training improves
the average score for all three model families: Qwen3-VL-4B from
$52.61$ to $55.23$, Qwen2.5-VL-7B from $40.85$ to $43.35$, and
MiMo-VL-7B-SFT from $63.37$ to $66.50$.  The backbones differ in
pretraining recipe, generation style, and starting strength, so the
consistent gains are unlikely to be an artifact of one family or
one evaluation format.

The gains are not concentrated in one benchmark type, evidence that
environment diversity matters.  \tron{} improves each backbone
across multiple benchmark families covering symbolic and geometric
calculation, spatial reasoning, logical inference, and algorithmic
visual puzzles, suggesting that the environments transfer through
practiced operations rather than memorized benchmark templates.

The strongest relative pattern is on structured-reasoning
benchmarks: MM-HELIX improves for every backbone, and SpatialEval
gains substantially for both Qwen2.5-VL-7B and MiMo-VL-7B-SFT.
These tasks align with the mechanisms \tron{} emphasizes
(deterministic state transitions, grid or graph structure,
geometric constraints, exact answer checking).  Not every column
moves equally; the ability-specialist analysis below shows that
transfer is better explained by the underlying capability than by
the benchmark category.

MiMo-VL-7B-SFT is the strongest starting point yet gains the most
in mean score, suggesting that rule-verifiable RL contributes
signal complementary to supervised long-reasoning tuning.

\subsection{Ability-Specialist Results}
\label{sec:ability_specialists}

\paragraph{Visual format vs.\ underlying capability.}
To analyze the specialist results, we first separate two
properties of a VQA problem.  Its \emph{visual format} is the
surface category of the image, identifiable without solving the
task: charts and bar plots fall under ``chart/diagram'',
line drawings with labelled angles and segments under
``math'', and so on.  Its \emph{underlying capability} is the
computation a correct solver must actually perform on that
image to produce the answer.  The underlying capability can
diverge from what the format suggests: a geometry problem may
rely more on value reading than on theorem chaining, and a chart
question may rely more on multi-step inference than on chart
parsing.  Both \tron{} buckets and external
benchmark subtasks are classified primarily by visual format,
so any specialist analysis must distinguish that
format axis from the capability the problem actually demands.
Each \tron{} environment is constructed around a single core
ability, whereas a real evaluation task often combines multiple
capabilities.

We organize the analysis around three questions:
\textbf{RQ1:} Can a visual-format-defined bucket effectively
train its specialist on subtasks within its domain?
\textbf{RQ2:} Does the underlying capability transfer across visual
formats?
\textbf{RQ3:} Is visual-format alignment alone sufficient when the
underlying capability does not match?

Each specialist starts from Qwen3-VL-4B-Instruct and uses the DAPO RL recipe, but samples from only one bucket: Math, Spatial,
Count, Pattern, or Diagram.
Appendix~\ref{app:ability_training_details} details the setup.

\smallskip{}

\FloatBarrier

\begin{table}[!t]
\centering
\scriptsize
\setlength{\tabcolsep}{2.0pt}
\renewcommand{\arraystretch}{0.98}
\begin{tabular}{@{}l>{\raggedright\arraybackslash}m{0.24\columnwidth}>{\raggedright\arraybackslash}m{0.30\columnwidth}cc@{}}
\toprule
\textbf{Spec.} & \textbf{Practiced ability} & \textbf{Benchmark/subtask}
& \textbf{Score} & $\Delta$ \\
\midrule
\multirow{3}{*}{Math}
& Visual geometry & WeMath angles/length & 39.8$\rightarrow$\textbf{51.0} & \dnum{+11.2} \\
& Coord.\ geometry & WeMath coord./pos. & 76.7$\rightarrow$\textbf{83.3} & \dnum{+6.7} \\
& Arith.\ puzzle & MM-HELIX 24Points & 83.3$\rightarrow$\textbf{90.0} & \dnum{+6.7} \\
\midrule
\multirow{3}{*}{Spatial}
& Path planning & MM-HELIX maze & 36.7$\rightarrow$\textbf{53.3} & \dnum{+16.7} \\
& Position reason. & WeMath position & 63.1$\rightarrow$\textbf{73.3} & \dnum{+10.3} \\
& Spatial inference & LogicVista spatial & 20.5$\rightarrow$\textbf{29.5} & \dnum{+9.0} \\
\midrule
\multirow{3}{*}{Count}
& Extrema count & MM-HELIX hills/val.\ & 26.7$\rightarrow$\textbf{36.7} & \dnum{+10.0} \\
& Chart numeracy & CharXiv num.-in-chart & 44.8$\rightarrow$\textbf{47.8} & \dnum{+3.0} \\
& General numeracy & CharXiv num.-in-gen.\ & 31.9$\rightarrow$\textbf{33.2} & \dnum{+1.3} \\
\midrule
\multirow{3}{*}{Pattern}
& Constraint puzzle & MM-HELIX tapa & 13.3$\rightarrow$\textbf{20.0} & \dnum{+6.7} \\
& Pattern analysis & MME-R pattern & 38.2$\rightarrow$\textbf{40.3} & \dnum{+2.1} \\
& Inductive reason. & LogicVista inductive & 29.9$\rightarrow$\textbf{31.8} & \dnum{+1.9} \\
\midrule
\multirow{3}{*}{Diagram}
& Chart text read & CharXiv text-in-chart & 39.1$\rightarrow$\textbf{41.8} & \dnum{+2.7} \\
& Stat.\ visuals & DynaMath statistics & 72.2$\rightarrow$\textbf{74.6} & \dnum{+2.5} \\
& Hypothet.\ chart & ChartQAPro hypothet.\ & 37.3$\rightarrow$\textbf{39.3} & \dnum{+2.0} \\
\bottomrule
\end{tabular}
\caption{Within-bucket gains: each specialist on external benchmark
subtasks whose visual format matches its training bucket. $\Delta$ is Spec.\ minus Base.}
\label{tab:ability_signatures}
\end{table}

\begin{table}[!t]
\centering
\scriptsize
\setlength{\tabcolsep}{2.0pt}
\renewcommand{\arraystretch}{0.98}
\begin{tabular}{@{}l>{\raggedright\arraybackslash}m{0.24\columnwidth}>{\raggedright\arraybackslash}m{0.30\columnwidth}cc@{}}
\toprule
\textbf{Spec.} & \textbf{Adjacent ability} & \textbf{Benchmark/subtask}
& \textbf{Score} & $\Delta$ \\
\midrule
\multirow{3}{*}{Math}
& Grid traversal & MM-HELIX maze & 36.7$\rightarrow$\textbf{56.7} & \dnum{+20.0} \\
& Coord.\ position & WeMath position & 63.1$\rightarrow$\textbf{74.7} & \dnum{+11.6} \\
& Spatial inference & LogicVista spatial & 20.5$\rightarrow$\textbf{32.1} & \dnum{+11.5} \\
\midrule
\multirow{3}{*}{Spatial}
& Angle measure & WeMath angles/length & 39.8$\rightarrow$\textbf{52.5} & \dnum{+12.6} \\
& Map navigation & WeMath route map & 66.8$\rightarrow$\textbf{73.9} & \dnum{+7.1} \\
& 3D volume & MathVerse volume & 32.2$\rightarrow$\textbf{37.3} & \dnum{+5.1} \\
\midrule
\multirow{3}{*}{Count}
& Direction sense & WeMath direction & 89.5$\rightarrow$\textbf{100.0} & \dnum{+10.5} \\
& Unit 3D volume & MathVerse volume & 32.2$\rightarrow$\textbf{40.0} & \dnum{+7.8} \\
& Graph reasoning & DynaMath graph theory & 59.6$\rightarrow$\textbf{63.7} & \dnum{+4.2} \\
\midrule
\multirow{3}{*}{Pattern}
& Path constraint & MM-HELIX Hamilton & 20.0$\rightarrow$\textbf{26.7} & \dnum{+6.7} \\
& Cyclic pattern & PuzzleVQA size cycle & 26.0$\rightarrow$\textbf{32.0} & \dnum{+6.0} \\
& Chart numeracy & CharXiv num.-in-chart & 44.8$\rightarrow$\textbf{48.3} & \dnum{+3.4} \\
\midrule
\multirow{3}{*}{Diagram}
& Numeric pattern & PuzzleVQA rect-height & 31.0$\rightarrow$\textbf{41.0} & \dnum{+10.0} \\
& Visual numeracy & CharXiv num.-in-gen.\ & 31.9$\rightarrow$\textbf{38.9} & \dnum{+7.0} \\
& Struct.\ induction & LogicVista inductive & 29.9$\rightarrow$\textbf{35.5} & \dnum{+5.6} \\
\bottomrule
\end{tabular}
\caption{Cross-bucket gains: each specialist on external benchmark
subtasks whose visual format lies outside its training bucket.
$\Delta$ is Spec.\ minus Base.}
\label{tab:ability_cross_transfer}
\end{table}

\begin{table*}[!t]
\centering
\scriptsize
\setlength{\tabcolsep}{2.1pt}
\renewcommand{\arraystretch}{1.12}
\makebox[\textwidth][c]{%
\resizebox{\textwidth}{!}{%
\begin{tabular}{@{}l@{\hspace{5pt}}rrrr@{\hspace{6pt}}rrrr@{\hspace{6pt}}rrr@{\hspace{5pt}}r@{}}
\toprule
\textbf{Run}
& \multicolumn{4}{c}{\textbf{Mathematical reasoning}}
& \multicolumn{4}{c}{\textbf{Spatial and logical reasoning}}
& \multicolumn{3}{c}{\textbf{Charts, figures, puzzles}}
& \textbf{Mean} \\
\cmidrule(lr){2-5} \cmidrule(lr){6-9} \cmidrule(lr){10-12} \cmidrule(l){13-13}
& \textbf{WeMath-S} & \textbf{WeMath-L}
& \textbf{MathV} & \textbf{Dyna}
& \textbf{MME-R} & \textbf{Spat.} & \textbf{Logic} & \textbf{HELIX}
& \textbf{CharXiv} & \textbf{ChartQA} & \textbf{Puzzle}
& \\
\midrule
Base
& 52.86 & 70.95 & 43.15 & 66.11
& 42.85 & 77.52 & 57.05 & 21.67
& 39.80 & 34.41 & \underline{72.35} & 52.61 \\
\midrule
Math
& \textbf{58.76} & \textbf{75.90} & 41.22 & 66.25
& 44.70 & 77.61 & 59.51 & \textbf{26.90}
& \underline{42.00} & \textbf{35.52} & 70.45 & 54.44 \\
Spatial
& 56.38 & \underline{75.62} & 42.56 & 65.41
& 44.11 & 78.75 & 59.06 & 23.65
& 39.30 & \textbf{35.52} & 71.85 & 53.84 \\
Count
& 57.14 & 74.29 & 44.14 & 66.27
& \underline{45.20} & 77.52 & 59.28 & 23.97
& \underline{42.00} & 34.08 & 69.90 & 53.98 \\
Pattern
& 54.48 & 71.81 & 41.57 & 65.33
& 43.27 & 76.22 & 56.38 & 23.49
& 41.60 & 34.80 & 71.45 & 52.76 \\
Diagram
& 57.81 & 74.76 & \textbf{44.67} & \underline{66.69}
& 44.70 & \underline{79.01} & \textbf{60.18} & 25.40
& 41.90 & 33.94 & 71.40 & \underline{54.59} \\
\midrule
Full
& \underline{58.29} & 74.86 & \underline{44.54} & \textbf{67.49}
& \textbf{45.88} & \textbf{80.47} & \underline{59.73} & \underline{25.56}
& \textbf{42.40} & \underline{35.04} & \textbf{73.25} & \textbf{55.23} \\
\bottomrule
\end{tabular}
}%
}
\caption{Broad-suite effect of ability specialists.  Each specialist
trains on one ability bucket; Full is the jointly trained model.
Bold marks the best score per benchmark or mean, and underlining marks
the second-best score.
}
\label{tab:ability_specialists}
\end{table*}

\noindent\textbf{RQ1 (within-bucket alignment).}
Since each bucket is defined by a visual format, naturally the first
question is whether training on such a bucket actually improves
subtasks in external benchmarks that share the same visual format.
Table~\ref{tab:ability_signatures} answers this by reporting, for
each specialist, representative gains on external benchmark subtasks
within its bucket's domain: geometry subtasks for Math, path-planning
and position subtasks for Spatial, counting subtasks for Count, and
so on.  Each specialist substantially improves these subtasks: Math
gains $+11.2$ on WeMath angles/length, Spatial $+16.7$ on MM-HELIX
maze, Count $+10.0$ on MM-HELIX hills/valleys, with similar
improvements across all five buckets.
\underline{Takeaway}: visual-format alignment within a bucket has a
clear effect: each specialist substantially improves external
subtasks whose visual format matches its bucket.

\noindent\textbf{RQ2 (cross-bucket carryover).}
Table~\ref{tab:ability_cross_transfer} shows that each specialist
also lifts external subtasks outside its bucket, where the visual
format differs from its training but a shared underlying capability
remains.  The most dramatic case is Math on MM-HELIX maze
($+20.0$), alongside WeMath position ($+11.6$) and LogicVista
spatial ($+11.5$), consistent with Math training a transferable
multi-step reasoning capability rather than a maze-specific skill.
Spatial transfers to WeMath angles/length ($+12.6$) and route map
($+7.1$), consistent with a shared spatial-understanding capability
across formats.  Count transfers to MathVerse volume ($+7.8$) and
DynaMath graph theory ($+4.2$), consistent with a visual-grounding
capability over discrete elements.  Pattern transfers to MM-HELIX
Hamiltonian path ($+6.7$), consistent with constraint and rule
reasoning over discrete visual structures.  Diagram's bar-chart
training transfers to PuzzleVQA rect-height ($+10.0$), consistent
with a figure-reading capability.
\underline{Takeaway}: as long as the underlying capability a task
requires has been trained, that capability can transfer to the
task even when the task's visual format does not appear in
training.

\noindent\textbf{RQ3 (visual format alone).}
Table~\ref{tab:ability_specialists} reports each specialist's
accuracy across the 10 external benchmarks.  The visually-aligned
specialist (whose bucket shares the benchmark's visual format) wins
only on WeMath; two cases show why.  On MathVerse, three of the
five problem versions strip the textual description, so the
bottleneck is figure-reading across structured geometric visuals
rather than theorem chaining: Diagram trains figure-reading and
wins, while Math bundles figure-reading inside theorem chains where
it never becomes a standalone skill, and ends up regressing below
Base.  CharXiv (Reasoning split) is the 
opposite: a multi-step
reasoning load over scientific charts on which Diagram falls behind
Math because the questions require compositional parsing,
ordinal comparison, and chained inference; Math trains exactly these
dense inference chains, while Diagram's chart extraction and
aggregation are too shallow.
\underline{Takeaway}: both visual format and underlying capability
contribute to task improvement (RQ1 and RQ2), but visual-format
alignment alone is not sufficient: the visually-aligned specialist
wins only $1$ of $10$ benchmarks, so an effective training set
should cover both the format and the underlying capabilities a task
demands.


\section{Discussion}
\label{sec:discussion}

\noindent\textbf{Live environments preserve freshness and an
advancing curriculum.}
Parametric live environments give every sampled instance an exact
reward, supply fresh latent states and renderings that keep
memorization pressure low, and expose a per-environment difficulty
ladder the curriculum can advance on demand.  Pre-generating
\tron{} rollouts into a static parquet would forfeit the last two
properties: the snapshot is bounded and exhausts once seen, its
curriculum cannot advance after collection, and its bucket mix is
frozen at generation time, so ability-specialist re-targeting
becomes a re-collection effort rather than a sampler switch.

\noindent\textbf{Underlying capability is the broader driver of transfer.}
A benchmark's visual format describes its inputs, not the
underlying capabilities a model must exercise to solve it, and a
single benchmark typically demands several capabilities at once.
Our specialist analysis (Section~\ref{sec:ability_specialists})
shows that single-bucket RL transfers better along the capability axis
rather than the visual-format axis.

\noindent\textbf{Diversity hedges against unidentified
capabilities.}
In practice, the underlying capability of a real task is hard to
decouple: a single question often interleaves several
capabilities and has no canonical breakdown, so we cannot reliably
identify in advance which capability a future task will demand.
The simplest robust response is to make the training-environment
set as diverse as possible, covering as many underlying
capabilities as we can, so that whichever capability an unseen task
actually demands is likely already in the mix.  This is the
design rationale behind \tron{}'s 520 environments, spanning
five ability buckets and, within each, as many
visual formats, generation mechanisms, and parameter ranges as
we could audit.

\section{Conclusion}
\label{sec:conclusion}

We introduced \tron{}, an online environment substrate for visual
reasoning RL: 520 generator-verifier programs organized into five
ability buckets, each producing fresh image-question rollouts with
exact rule-based rewards and a local difficulty ladder. We pair
the substrate with an audit that measures quality, diversity, and
difficulty before training. Across Qwen3-VL-4B, Qwen2.5-VL-7B, and
MiMo-VL-7B-SFT, RL post-training with \method{} consistently
improves performance on ten external multimodal reasoning
benchmarks, supporting online environments as a practical substrate
for visual reasoning training.

\section{Limitations}
\label{sec:limitations}

\tron{} environments are synthetic, so their visual style and
language can diverge from real benchmark data; the audit catches
internal quality failures but cannot guarantee distributional
alignment with every external benchmark, especially for domains
needing photographic or dense scientific perception.

Difficulty levels are author-chosen generator parameters.  The
aggregate base-model pass rate falls monotonically across levels
(Figure~\ref{fig:audit_summary}c), but individual environments need
not be strictly monotone and step sizes can vary.

The diversity analysis depends on hand-chosen hyperparameters:
signal normalizations, per-component weights inside $D_s$ and
$D_l$, $\mathrm{dup}$ thresholds, and the convex-combination
weights $(w_s,w_l,w_x)$ with A--D grade cutoffs
(Appendix~\ref{app:diversity_details}).  These were chosen to make
the grade histogram informative on the current suite rather than
learned, and a different operational definition of ``sufficient
diversity'' would shift the boundaries.

The five ability buckets are not strictly decoupled: many
environments exercise more than one mechanism (a chart task may
also need multi-step reasoning, a graph-algorithm task may also
need counting), so the labels should be read as a coarse partition
rather than a clean factorization.

\section{Ethical Considerations}

\paragraph{Data and models.} All training environments in \tron{}
are generated procedurally; we do not collect or scrape any new
data, and no personally identifiable information is involved.
Evaluation uses public multimodal reasoning benchmarks
(Section~\ref{sec:benchmarks}) and open-source vision-language
models (Qwen3-VL-4B-Instruct, Qwen2.5-VL-7B-Instruct, and
MiMo-VL-7B-SFT) under their respective licenses.

\paragraph{Intended use and misuse.} Our work targets improving
the visual reasoning capabilities of vision-language models
through procedurally generated, rule-verifiable environments. It
is intended for benign applications such as educational tools,
assistive vision, scientific figure understanding, and automated
analysis of structured visuals. The improved reasoning ability
could in principle be repurposed for surveillance or other
sensitive monitoring tasks; however, the contribution here is
methodological (a training substrate and curriculum framework)
and is not tailored toward such use cases.

\paragraph{Use of LLMs.} Large language models were used for
language polishing (grammar and phrasing) and for coding
assistance, including writing and debugging portions of the
procedural environment generators and analysis code. All research
ideas, problem formulations, methodology, experimental design,
analyses, and claims are the authors' own.

\clearpage
\bibliography{custom}

\clearpage
\appendix

\section{Fine-Grained Environment Coverage}
\label{app:capability_map}

Table~\ref{tab:capability_map_full} expands the high-level suite
composition in Table~\ref{tab:buckets}.  The entries are representative
rather than exhaustive; each listed environment is a generator--verifier
program with multiple seeds and difficulty levels.

\begin{table*}[p]
\centering
\tiny
\setlength{\tabcolsep}{2pt}
\renewcommand{\arraystretch}{0.98}
\begin{tabularx}{\textwidth}{@{}p{0.13\textwidth}p{0.19\textwidth}X p{0.26\textwidth}@{}}
\toprule
\textbf{Bucket} & \textbf{Fine-grained domain}
& \textbf{Representative environments}
& \textbf{Reasoning operations covered} \\
\midrule
\multirow{5}{0.13\textwidth}{\raggedright Spatial Reasoning}
& 3D rotation and chirality
& \emph{spatial rotation}, \emph{polycube rotation axis identify},
\emph{shape rotation invariant}, \emph{chiral object identification},
\emph{chirality pair discrimination}
& Mental rotation, handedness, invariant matching, and axis-based
object transformation. \\
& Projection, folding, and cross-sections
& \emph{net folding}, \emph{orthographic projection},
\emph{projection view}, \emph{solid cross section reverse},
\emph{unfold path prediction}
& Mapping between 2D views and 3D structure; inferring hidden geometry
from folds, nets, and slices. \\
& Navigation and relative direction
& \emph{map distance}, \emph{maze solution length},
\emph{relative direction chain}, \emph{bearing compass},
\emph{map route optimization}
& Route planning, shortest paths, compass direction, egocentric
reference frames, and chained spatial relations. \\
& Object layout and viewpoint
& \emph{before after}, \emph{depth order}, \emph{near far mcq},
\emph{perspective shift}, \emph{object tracking across frames}
& Viewpoint changes, occlusion order, temporal tracking, and relative
object placement. \\
& Mechanical and physical diagrams
& \emph{lever pulley}, \emph{gear rotation direction},
\emph{scale balance}, \emph{pendulum compare},
\emph{submersion water rise}
& Physical causality, force/ratio reasoning, mechanical linkage, and
qualitative dynamics. \\
\addlinespace[2pt]
\multirow{5}{0.13\textwidth}{\raggedright Mathematical Reasoning}
& Geometry theorem reasoning
& \emph{angle chain}, \emph{angle bisector chain deep},
\emph{circle theorem}, \emph{secant tangent},
\emph{cyclic quadrilateral advanced}, \emph{ptolemy quad}
& Multi-step angle chasing, circle facts, quadrilateral constraints,
and theorem-backed diagram reasoning. \\
& Coordinate and analytic geometry
& \emph{coordinate geometry}, \emph{analytic geom chain},
\emph{distance between points}, \emph{line parallel perp},
\emph{two lines intersection}, \emph{conic eccentricity mcq}
& Reading coordinates, applying formulas, comparing slopes, solving
intersections, and identifying conic properties. \\
& Functions and calculus from graphs
& \emph{function graph value read}, \emph{derivative graph},
\emph{parabola vertex}, \emph{implicit function level set},
\emph{area under curve estimation}, \emph{continuity at point}
& Graph-to-symbol translation, local/global function properties,
rate-of-change, and visual calculus. \\
& Probability and statistics
& \emph{probability tree}, \emph{spinner probability},
\emph{conditional probability visual}, \emph{histogram},
\emph{box plot comparison}, \emph{confusion matrix diagonal}
& Event composition, conditional probability, distribution reading,
summary statistics, and diagnostic-table reasoning. \\
& Algebraic and arithmetic visuals
& \emph{expression substitute evaluate}, \emph{matrix operation},
\emph{omitted operator}, \emph{sliding sum blanks},
\emph{numeric commonsense visual}, \emph{unit conversion visual}
& Symbolic substitution, matrix arithmetic, missing-operator search,
numeric commonsense, and unit conversion. \\
\addlinespace[2pt]
\multirow{5}{0.13\textwidth}{\raggedright Visual Diagram Understanding}
& Chart extraction and aggregation
& \emph{chart multistep}, \emph{bar chart aggregate},
\emph{chart aggregate claim}, \emph{chart filter aggregate},
\emph{chart kth largest}, \emph{chart percent change}
& Value lookup, aggregation, ranking, filtering, percent change, and
claim verification over charts. \\
& Complex plots and scientific figures
& \emph{dual axis chart}, \emph{ternary plot},
\emph{heatmap pattern identification}, \emph{phase diagram},
\emph{scientific graph interpretation}, \emph{kinematics graph}
& Multi-axis reading, scientific plot interpretation, phase/region
reasoning, and trend or threshold inference. \\
& Tables, schedules, and joins
& \emph{table cell lookup}, \emph{business table},
\emph{pivot table}, \emph{multi table join},
\emph{schedule table}, \emph{boarding pass duration}
& Cell lookup, row/column aggregation, relational joins, schedule
arithmetic, and structured-record comparison. \\
& Flow, graph, and process diagrams
& \emph{flowchart}, \emph{process flow diagram},
\emph{dependency graph}, \emph{circuit output prediction},
\emph{food web}, \emph{labeled parts diagram}
& Following arrows and dependencies, circuit logic, process-state
updates, and label-to-part grounding. \\
& Text, labels, and infographic reasoning
& \emph{text render math}, \emph{handwritten expression},
\emph{chart with latex label}, \emph{chart with context paragraph},
\emph{infographic business}, \emph{ambiguous label resolution}
& OCR-like reading, label disambiguation, text-plus-visual grounding,
and paragraph-conditioned visual reasoning. \\
\addlinespace[2pt]
\multirow{5}{0.13\textwidth}{\raggedright Visual Pattern \& Logical Reasoning}
& Visual analogy and sequences
& \emph{visual analogy raven 3x3}, \emph{analogy multiple dimensions},
\emph{analogy from sequence}, \emph{figure sequence next},
\emph{visual sequence}, \emph{matrix completion 5x5}
& Raven-style abstraction, analogy transfer, sequence continuation,
and matrix completion. \\
& Rule induction and classification
& \emph{rule induction sequence}, \emph{pattern rule multi example},
\emph{inductive rule discovery}, \emph{visual rule exception},
\emph{attribute enumeration discovery}, \emph{odd one out}
& Inferring latent rules, handling exceptions, attribute selection, and
category comparison. \\
& Deductive and symbolic logic
& \emph{logic grid}, \emph{syllogism passage}, \emph{quantifier logic},
\emph{truth table 3variable}, \emph{logical negation chain},
\emph{argument contradiction}
& Constraint propagation, syllogistic reasoning, truth tables,
quantifiers, negation, and contradiction detection. \\
& Constraint puzzles
& \emph{calcudoku}, \emph{futoshiki solve}, \emph{hitori solve},
\emph{binairo solve}, \emph{nonogram solve},
\emph{skyscrapers solve}, \emph{numbrix solve}
& Grid constraints, local/global consistency, search, and exact puzzle
verification. \\
& Algorithmic and graph search
& \emph{eulerian path find}, \emph{hamiltonian path find},
\emph{kruskal first edge}, \emph{topological sort},
\emph{shortest path directed weighted}, \emph{stack queue trace}
& Graph traversal, path existence, algorithm tracing, data-structure
state updates, and weighted shortest paths. \\
\addlinespace[2pt]
\multirow{5}{0.13\textwidth}{\raggedright Counting \& Quantitative Estimation}
& Object and instance counting
& \emph{object count with occlusion}, \emph{precise counting},
\emph{visual counting ultra easy}, \emph{shape instance count},
\emph{dense 2d count warmup}, \emph{missing grid count}
& Counting visible and missing items, handling distractors, and
separating object instances from background clutter. \\
& Region, cell, and path counts
& \emph{region counting}, \emph{region count grid},
\emph{grid cell count with rules}, \emph{path counting},
\emph{path length grid count}, \emph{lattice path count}
& Counting cells, regions, admissible paths, path lengths, and
rule-conditioned grid structures. \\
& 3D and stacked counts
& \emph{isometric counting}, \emph{multi view cube count},
\emph{hidden cube inference}, \emph{layered stack count},
\emph{cube decomposition count}, \emph{polycube counting warmup}
& Inferring counts from 3D views, hidden blocks, layers, decompositions,
and polycube structure. \\
& Attribute grouping and quantization
& \emph{attribute count quantize}, \emph{attribute height quantize},
\emph{attribute ordering}, \emph{feature counting classification},
\emph{stroke continuity grouping}, \emph{shape symmetry grouping}
& Grouping by visual attributes, quantizing continuous features,
ordering, and classifying by shape or stroke cues. \\
& Robust perception and traps
& \emph{blank annotation trap}, \emph{chart no labels},
\emph{chart unanswerable trap}, \emph{mcq answer not in options},
\emph{optical illusion lines}, \emph{visual occlusion counting}
& Detecting missing evidence, rejecting invalid options, handling
unanswerable prompts, illusions, and occlusion. \\
\bottomrule
\end{tabularx}
\caption{Fine-grained capability map for the \tron{} environment
suite.  This table expands Table~\ref{tab:buckets} by listing the
subdomains that drive environment authoring and representative
generator--verifier programs used during training.}
\label{tab:capability_map_full}
\end{table*}

\section{Qualitative Environment Examples}
\label{app:qual_examples}

\makeatletter
\setlength{\@fptop}{0pt}
\setlength{\@dblfptop}{0pt}
\makeatother

\begin{figure*}[p]
\centering
\includegraphics[width=\textwidth]{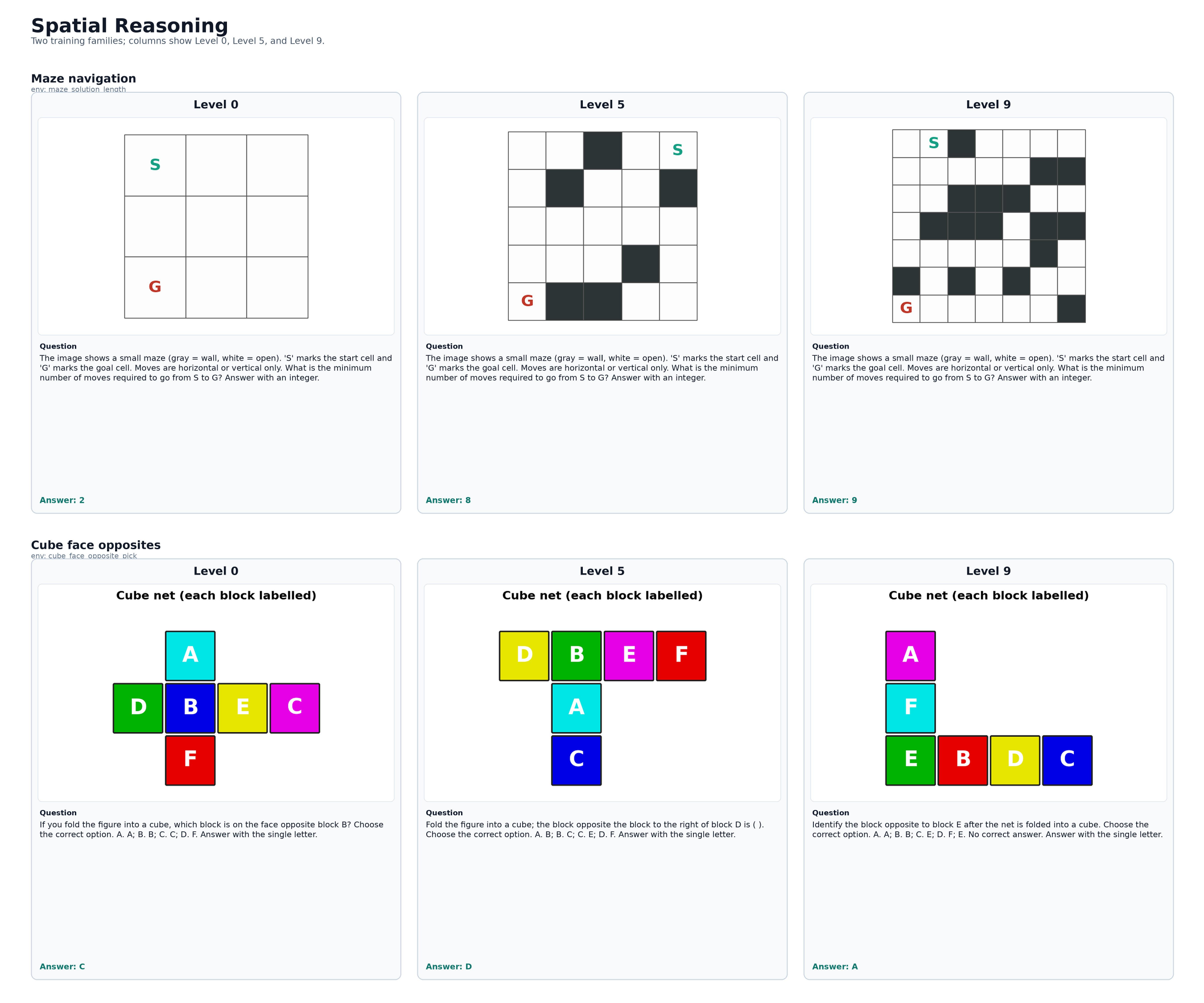}
\caption{Spatial Reasoning examples.  Rows show maze navigation and
cube-net opposite-face reasoning; columns increase the difficulty
level while keeping the generator family fixed.}
\label{fig:examples_spatial}
\end{figure*}

\begin{figure*}[p]
\centering
\includegraphics[width=\textwidth]{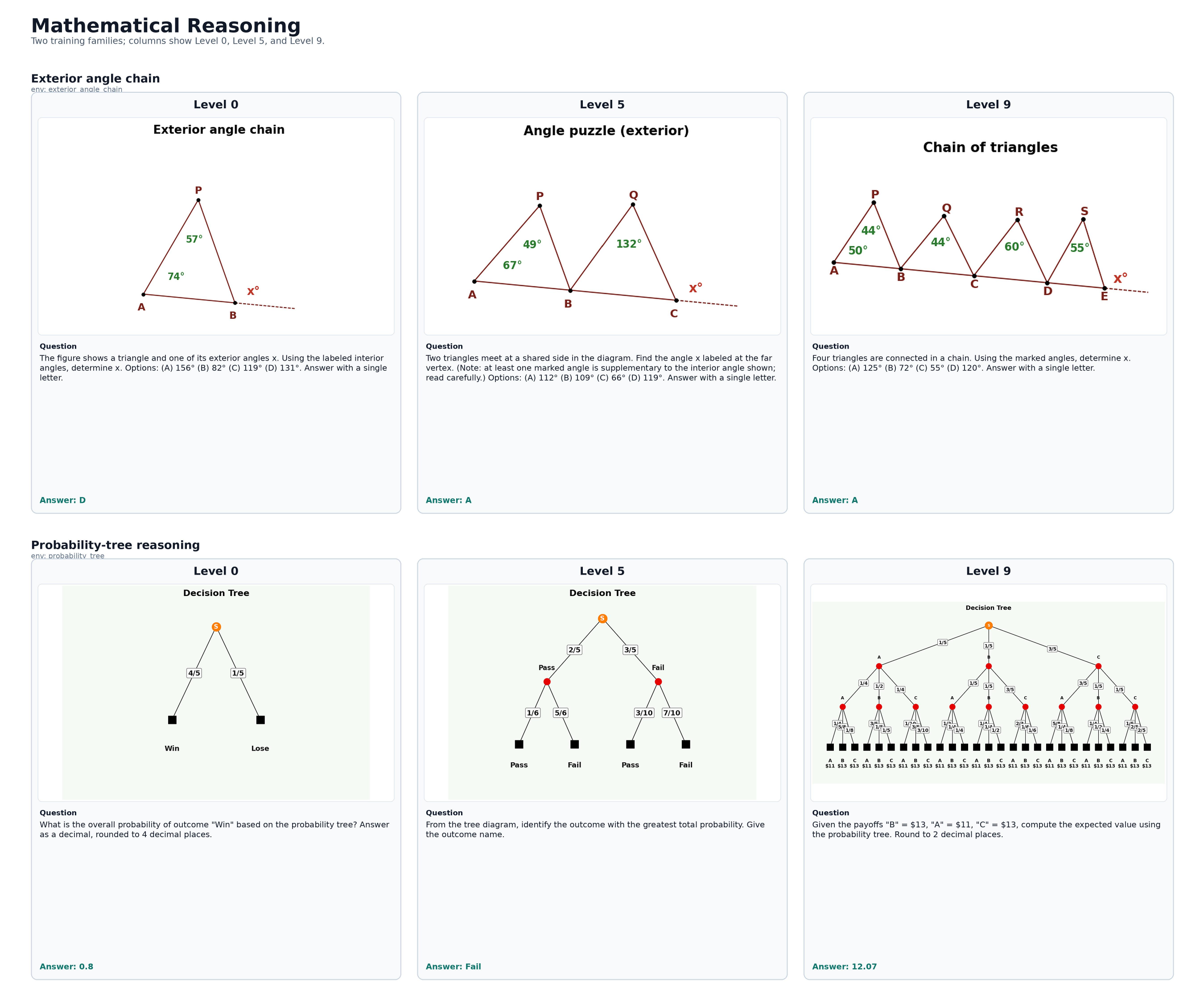}
\caption{Mathematical Reasoning examples.  Rows show exterior-angle
geometry and probability-tree reasoning.  Difficulty increases through
longer angle chains, denser trees, and more compositional numerical
queries.}
\label{fig:examples_math}
\end{figure*}

\begin{figure*}[p]
\centering
\includegraphics[width=\textwidth]{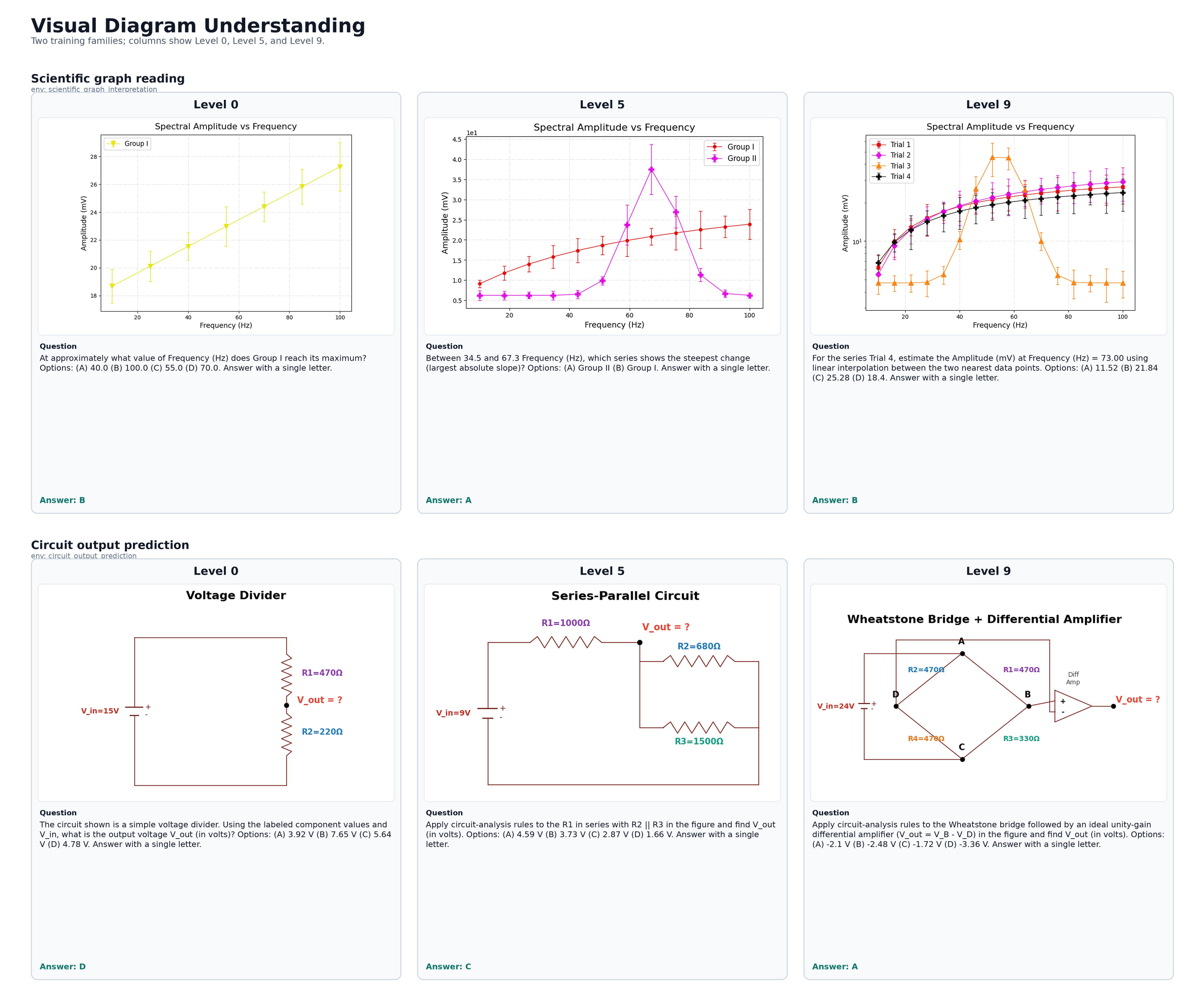}
\caption{Visual Diagram Understanding examples.  Rows show scientific
graph interpretation and circuit output prediction.  Higher levels add
more plotted series, interpolation, and more complex circuit topology.}
\label{fig:examples_diagram}
\end{figure*}

\begin{figure*}[p]
\centering
\includegraphics[width=\textwidth]{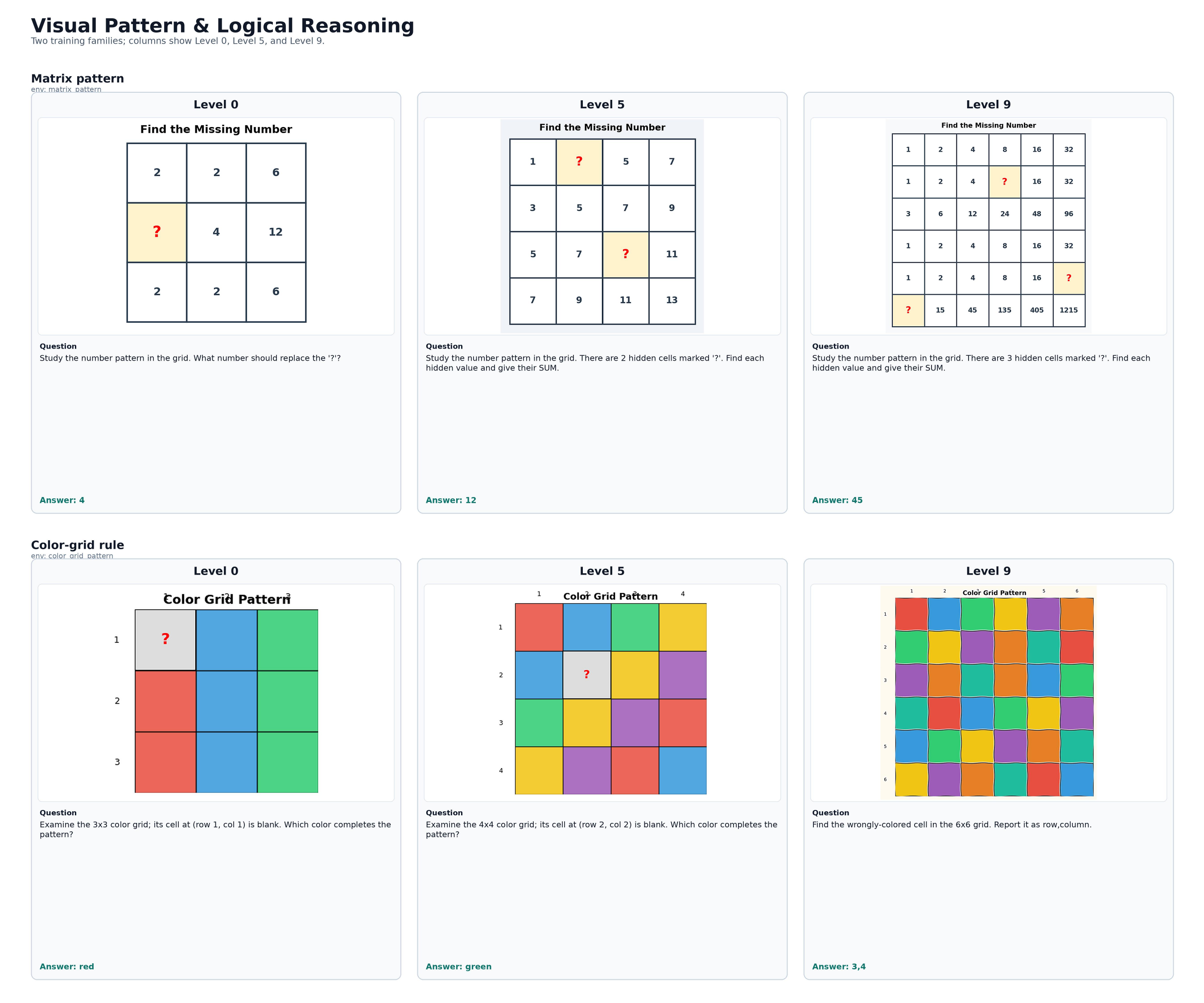}
\caption{Visual Pattern \& Logical Reasoning examples.  Rows show
matrix pattern completion and color-grid rule induction.  Higher levels
use larger grids, more symbols or colors, and harder rule violations.}
\label{fig:examples_pattern}
\end{figure*}

\begin{figure*}[p]
\centering
\includegraphics[width=\textwidth]{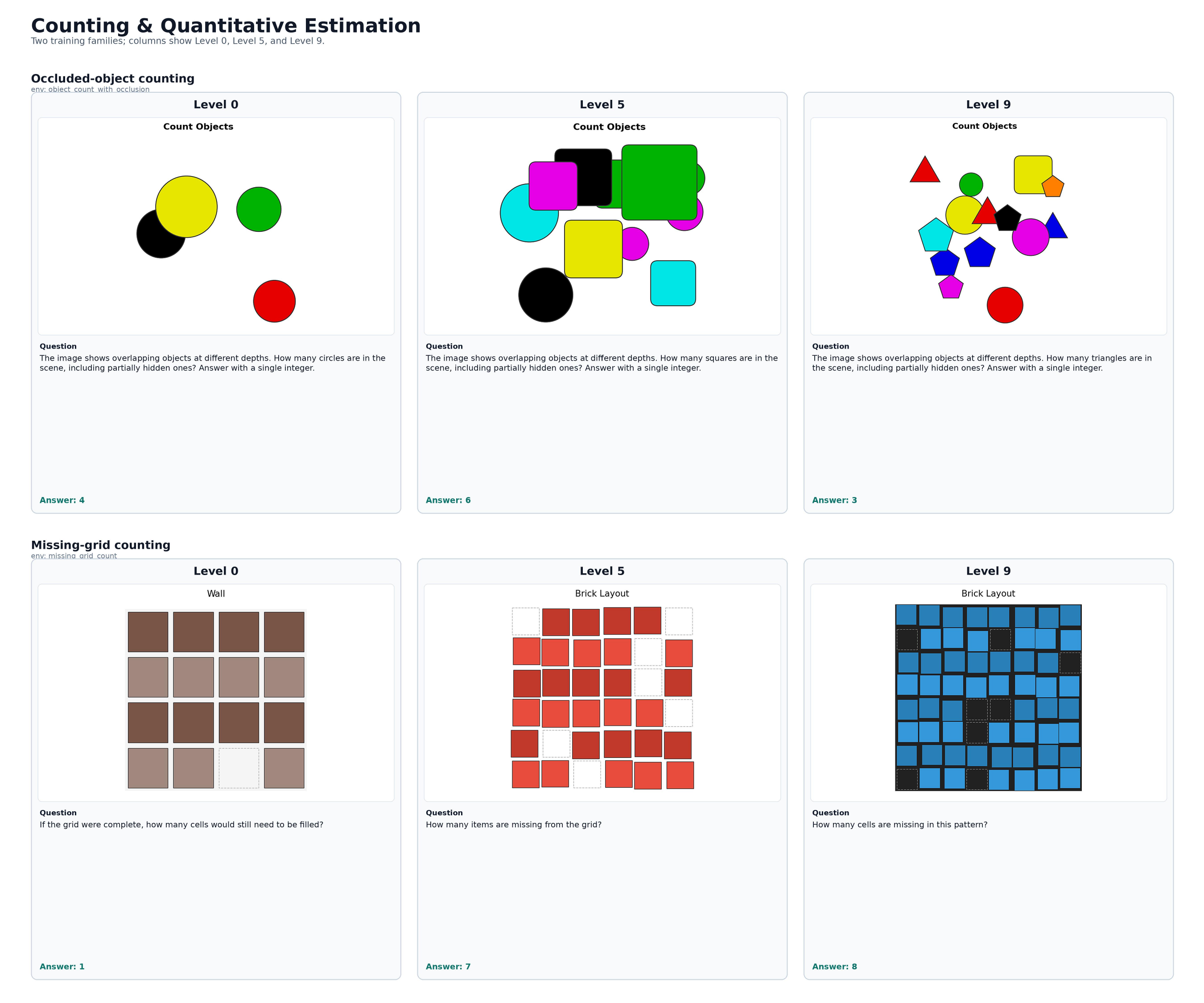}
\caption{Counting \& Quantitative Estimation examples.  Rows show
occluded-object counting and missing-grid counting.  Higher levels
increase clutter, occlusion, grid size, and the number of missing or
empty cells.}
\label{fig:examples_counting}
\end{figure*}

Figures~\ref{fig:examples_spatial}--\ref{fig:examples_counting}
show qualitative examples sampled directly from the 520 training
environments.  Each page focuses on one ability bucket, with two
generator families as rows and Levels 0, 5, and 9 as columns.  Each
panel pairs the rendered instance with its task prompt and verified
answer; only repeated answer-format boilerplate is omitted for
readability.  The examples therefore show both sides of the environment
contract: the visual instance given to the policy and the answer
accepted by the executable verifier.

\section{Full-Model Training Details}
\label{app:full_training_details}

The three full \tron{} runs reported in Table~\ref{tab:main_results}
share the same DAPO-style training recipe, the same online environment
sampler, and the same curriculum-promotion mechanism.  Backbone-specific
batch sizes and vLLM memory settings are adjusted to fit a four-GPU
H100 80\,GB node.  Table~\ref{tab:full_training_details} lists the
per-backbone hyperparameters, with shared settings repeated across
the three columns for readability.

\begin{table*}[h]
\centering
\small
\setlength{\tabcolsep}{6pt}
\renewcommand{\arraystretch}{1.1}
\begin{tabular}{@{}lccc@{}}
\toprule
\textbf{Setting} & \textbf{Qwen3-VL-4B} & \textbf{Qwen2.5-VL-7B} & \textbf{MiMo-VL-7B-SFT} \\
\midrule
Backbone HF weights & \texttt{Qwen3-VL-4B-Instruct} & \texttt{Qwen2.5-VL-7B-Instruct} & \texttt{MiMo-VL-7B-SFT} \\
Reported checkpoint step & 330 & 550 & 550 \\
Train batch size & 64 & 32 & 32 \\
PPO mini-batch size & 64 & 32 & 32 \\
PPO micro-batch / GPU & 2 & 2 & 2 \\
Log-prob micro-batch / GPU & 2 & 1 & 1 \\
\midrule
Rollouts per prompt ($n$) & 8 & 8 & 8 \\
Rollout temperature & 1.0 & 1.0 & 1.0 \\
Max prompt length & 8192 & 8192 & 8192 \\
Max response length & 8192 & 8192 & 8192 \\
Optimizer & \multicolumn{3}{c}{Adam, lr $5\!\times\!10^{-6}$, betas $(0.9, 0.98)$} \\
Clip ratio (low, high) & \multicolumn{3}{c}{(0.2,\ 0.28)} \\
Entropy coefficient & \multicolumn{3}{c}{0} \\
KL loss coefficient & \multicolumn{3}{c}{0.005 (low-var KL); not added to reward} \\
Group filter & \multicolumn{3}{c}{enabled (metric = accuracy, max gen batches = 10)} \\
Tensor parallelism & \multicolumn{3}{c}{4} \\
Hardware & \multicolumn{3}{c}{4 $\times$ H100 80\,GB} \\
\bottomrule
\end{tabular}
\caption{Full-model training settings for the three backbones
reported in the main results.  The upper block lists backbone-specific
differences (batch sizes scaled to fit 4 H100 GPUs, and the training
step at which the reported checkpoint was taken); the lower block is
identical across runs.  For each backbone the reported checkpoint is
selected at convergence on a held-out \tron{} validation parquet
(Appendix~\ref{app:ability_training_details}): the 4B run trains for
330 steps and the 7B and MiMo-VL runs train for 550 steps.}
\label{tab:full_training_details}
\end{table*}

\paragraph{Curriculum promotion rule (identical across all three runs).}
At each training step, every prompt generates $n=8$ rollouts that the
verifier scores 0/1.  The per-environment curriculum manager maintains
a sliding window of the four most recent rollout groups for the
environment's current level $\ell$, giving a buffer of
$4\times 8 = 32$ scored trajectories at any time.  Once the mean
accuracy over the buffer reaches $\geq 0.80$, the environment's level
is advanced from $\ell$ to $\ell+1$ and the buffer is reset.  Group
filtering removes uninformative prompt groups (all-correct or
all-wrong) from the policy update, but every scored rollout still
contributes to the curriculum accumulator.  The sampler keeps a
$0.30$ probability of mixing lower-level instances so that training
does not collapse onto only the hardest level reached.  Checkpoints
and validation parquets are saved every $10$ training steps.

\section{Ability-Specialist Training Details}
\label{app:ability_training_details}

The ability-specialist runs use the same DAPO/GRPO training stack as
the broad mixed run, but restrict the online environment sampler to one
ability bucket.  The launcher reads the bucket list, sets the
environment filter accordingly, and writes separate checkpoints,
validation generations, and curriculum state for each ability.  All
specialists start from Qwen3-VL-4B-Instruct and use the same
rule-based reward function as the full model.

\begin{table}[h]
\centering
\scriptsize
\setlength{\tabcolsep}{3pt}
\begin{tabular}{@{}lrrp{0.39\columnwidth}@{}}
\toprule
\textbf{Ability} & \textbf{Envs} & \textbf{Step} & \textbf{Comment} \\
\midrule
Math & 131 & 200 & mathematical reasoning \\
Spatial & 111 & 200 & spatial reasoning \\
Diagram & 144 & 200 & diagram and structured visual reasoning \\
Pattern & 104 & 200 & visual pattern and logic \\
Count & 30 & 100 & capped because the bucket is small \\
\bottomrule
\end{tabular}
\caption{Ability-specialist bucket sizes and reported checkpoints.}
\label{tab:ability_training_details}
\end{table}

Count is capped at step 100 because it has only 30 training
environments, compared with 104--144 environments for the other
buckets.  Running it to the same 200-step horizon would give the
Count specialist disproportionately many gradient updates per
environment, increasing the risk of overfitting to the small
bucket; the 100-step cap keeps the per-environment update count
roughly comparable to the other specialists.

For all specialists, the online training epoch size is 3200 generated
prompts.  The training batch, generation batch, and PPO mini-batch are
all 32, with eight rollouts per prompt.  Rollouts use temperature 1.0,
maximum prompt length 8192, and maximum response length 8192.  Training
uses four GPUs with vLLM tensor parallelism 4.  The actor learning rate
is \(5\times10^{-6}\), Adam betas are \((0.9,0.98)\), entropy
coefficient is 0, clipping uses the \([0.2,0.28]\) range, and the actor
KL loss coefficient is 0.005.  KL is not added directly to the reward.
Group filtering is enabled with accuracy as the filtering metric and at
most 10 generated batches per update; all generated rollouts are scored
before filtering and therefore remain available for curriculum
promotion.  Checkpoints and validation are run every 10 training steps.

The curriculum state is maintained per ability.  Promotion uses a
minimum accuracy threshold of 0.80, at least eight samples, eight
rollouts per prompt, and a difficulty-check batch of 16.  The sampler
keeps a 0.30 probability of mixing lower-level instances so that
training does not immediately collapse onto only the hardest level.
Each run reserves a fresh seed block at startup to avoid seed reuse
after crashes or restarts.

Each ability has a deterministic validation parquet generated from up
to 30 environments from the same bucket.  Validation samples levels
\(\{0,3,6,9\}\) with 20 seeds per level.  The auto-restart wrapper runs
validation before training on the first attempt to capture the step-0
baseline, then skips repeated baseline validation on restarts and
resumes from the latest checkpoint and curriculum snapshot.

\begin{figure}[h]
\centering
\includegraphics[width=\columnwidth]{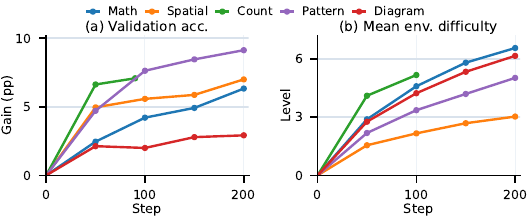}
\caption{Training dynamics for ability specialists.  The left panel
shows validation-accuracy gain over step 0; the right panel shows mean
curriculum difficulty.  Count stops at step 100; other specialists are
shown through step 200.}
\label{fig:ability_specialist_dynamics}
\end{figure}

Figure~\ref{fig:ability_specialist_dynamics} shows the resulting
training dynamics.  The left panel plots validation-accuracy gain
over the step-0 baseline; the right panel plots the mean
curriculum difficulty across audited levels.  All five
specialists improve monotonically on their bucket validation set
and advance their per-environment curriculum upward as lower
levels are mastered.  The Count curve stops at step 100 (per the
overfitting cap above), while the other specialists run through
step 200.

\section{Quality Gate Implementation}
\label{app:quality_gate_impl}

This appendix spells out the binary predicates used in
Section~\ref{sec:audit} (Quality paragraph), summarized in
Table~\ref{tab:quality_gate_impl}.  A requested probe is counted
in \(\mathcal O_e\) only if the environment generation call
completes and returns success; the remaining three gates are
evaluated over those successful probes.

The four gates target distinct failure modes.  The generation
gate catches generator-side exceptions and missing returns.  The
image gate catches blank, single-color, or saturated renderings
through size (\(\geq 64\)\,px), contrast (grayscale std
\(\geq 2.0\)), and foreground-ratio bounds
(\([0.001, 0.98]\) relative to the page median).  The
question/answer gate catches missing or empty fields after
normalization.  The verifier gate catches verifiers that accept
arbitrary strings or reject the canonical correct answer, by
checking both a wrapped correct payload (must score 1.0) and a
fixed wrong payload (must score 0.0).  All thresholds are coarse
syntactic bounds chosen to flag obvious failure modes without
penalizing environments with legitimately sparse or dense
images.

\begin{table}[h]
\centering
\small
\setlength{\tabcolsep}{3pt}
\begin{tabularx}{\columnwidth}{@{}lX@{}}
\toprule
\textbf{Gate} & \textbf{Predicate used by the audit} \\
\midrule
Generation &
The environment call succeeds without exception and returns a rendered
instance. \\
Image &
The rendered image has width and height at least 64 pixels, grayscale
standard deviation at least 2.0, and foreground-ratio proxy in
\([0.001,0.98]\).  The proxy counts pixels whose grayscale value differs
from the page median by more than 10. \\
Question/answer &
The normalized question string and ground-truth answer string are both
non-empty. \\
Verifier &
The verifier assigns full accuracy to a wrapped correct answer and zero
accuracy to a fixed wrong payload.  The audit rotates among the accepted
answer wrappers used during training. \\
\bottomrule
\end{tabularx}
\caption{Implementation-level predicates for the quality gates.  These
checks are deliberately syntactic and model-free; they catch broken
rendering, missing fields, and verifier failures before RL training.}
\label{tab:quality_gate_impl}
\end{table}

\section{Diversity Audit Details}
\label{app:diversity_details}

This appendix gives the concrete definitions of every symbol used in
Section~\ref{sec:audit}.

\paragraph{Setup.}
For each environment $e$, audited level $\ell\in\mathcal{L}$, and
successfully generated seed $k$, write
$x_{e,\ell,k}=(I_{e,\ell,k},q_{e,\ell,k},a_{e,\ell,k})$.  The audit
extracts three primitives from each probe: a 256-bit image perceptual
hash $\pi(x)$, a number-normalized question template $\tau(x)$, and an
answer bucket $\beta(x)$.  Let $K_\ell$ denote the number of
successful probes at level $\ell$ and $T_{e,\ell}$ the set of
templates seen.  Let $c(y)=\min(1,y)$ be the cap-at-1 function.

\paragraph{Seed signals $(h,t,a)$ at level $\ell$.}
These are the three signals fed to the seed-spread formula in
Section~\ref{sec:audit}:
\begin{align}
h_\ell &= c\!\left(\frac{3}{256}\,
  \operatorname*{avg}_{i<j} d_H(\pi(x_{e,\ell,i}),\pi(x_{e,\ell,j}))\right), \\
t_\ell &= \frac{|\{\tau(x_{e,\ell,k})\}_{k=1}^{K_\ell}|}
              {\max(1,K_\ell)}, \\
a_\ell &= \frac{H(\{\beta(x_{e,\ell,k})\}_{k=1}^{K_\ell})}
              {\log_2\max(2,K_\ell)},
\end{align}
where $d_H$ is Hamming distance and $H$ is Shannon entropy in bits.
Intuitively, $h_\ell$ measures how visually different the rendered
images are across seeds (large pairwise pHash distance);
$t_\ell$ measures whether the question wording varies (fraction
of distinct number-normalized templates); and $a_\ell$ measures
whether the answers vary (normalized entropy over answer
buckets).  A signal near $0$ indicates seed-collapse on that
axis.
The seed-spread weights are
$(w_h,w_t,w_a)=(0.45,\,0.25,\,0.30)$, giving the per-environment
seed-diversity score
\begin{equation}
D_s(e) = \operatorname*{avg}_{\ell\in\mathcal{L}}\,
  \bigl(w_h\,h_\ell + w_t\,t_\ell + w_a\,a_\ell\bigr).
\end{equation}
The factor $3$ in $h_\ell$ keeps small but real within-generator
pHash spread visible after the cap; without it, $h_\ell$
saturates near zero on most environments.

\paragraph{Level signals $(h,j,f)$ at pair $(\ell,\ell')$.}
For each adjacent pair $(\ell,\ell')\in\mathcal{A}$, with
$\bar f_{e,\ell}$ the mean foreground-pixel ratio at level $\ell$:
\begin{align}
h_{\ell,\ell'} &= c\!\left(\frac{1}{120}\,
  \operatorname*{avg}_{i,j} d_H(\pi(x_{e,\ell,i}),\pi(x_{e,\ell',j}))\right), \\
j_{\ell,\ell'} &= 1 - \frac{|T_{e,\ell}\cap T_{e,\ell'}|}
                          {|T_{e,\ell}\cup T_{e,\ell'}|}, \\
f_{\ell,\ell'} &= c\!\left(\frac{|\bar f_{e,\ell'}-\bar f_{e,\ell}|}
                                {0.20}\right).
\end{align}
Intuitively, $h_{\ell,\ell'}$ measures whether the rendered
images change between two difficulty levels (cross-level pHash
distance); $j_{\ell,\ell'}$ measures whether the question
templates turn over between levels (Jaccard distance over
template sets); and $f_{\ell,\ell'}$ measures whether the image
foreground complexity shifts between levels.  All three close to
$0$ would mean the difficulty axis changes only hidden metadata.
The level-shift weights are $(w_h,w_j,w_f)=(0.55,\,0.30,\,0.15)$,
giving the per-environment level-diversity score
\begin{equation}
D_l(e) = \operatorname*{avg}_{(\ell,\ell')\in\mathcal{A}}\,
  \bigl(w_h\,h_{\ell,\ell'} + w_j\,j_{\ell,\ell'}
        + w_f\,f_{\ell,\ell'}\bigr).
\end{equation}
The constants $120$ and $0.20$ are normalizers chosen so that a
typical between-level pHash distance and foreground-ratio change
both map into the $[0,1]$ range before clipping.  The $w_h$ in
$D_l$ is a different constant from the $w_h$ in $D_s$; both
multiply pHash-based signals, but in different formulas.

\paragraph{Cross-environment predicates $(C_\mathrm{hash},
C_\mathrm{thumb}, C_\mathrm{temp})$.}
For a pair of environments $(e,e')$, the audit aggregates their
$\ell=0$ probes into three cross-sample summaries: the mean pHash
Hamming distance $\bar d_H(e,e')$, the mean thumbnail-pixel mean
absolute error $\bar m(e,e')$, and the maximum token-Jaccard
similarity $J_{\max}(T_{e,0},T_{e',0})$ between any pair of
normalized L0 prompt templates from the two environments.  The three
predicates inside the $\mathrm{dup}$ formula of
Section~\ref{sec:audit} are
\begin{align}
C_\mathrm{hash}(e,e')  &= [\bar d_H(e,e') < 20], \\
C_\mathrm{thumb}(e,e') &= [\bar m(e,e') < 8], \\
C_\mathrm{temp}(e,e')  &= [J_{\max}(T_{e,0}, T_{e',0}) \geq 0.50].
\end{align}
Thresholds $(20,\,8,\,0.50)$ are chosen so that a flag requires
visual, pixel, and prompt similarity simultaneously.  The
per-environment indicator is $D_x(e)=0$ if there exists $e'\neq e$
with $\mathrm{dup}(e,e')=1$ and $D_x(e)=1$ otherwise.

\paragraph{Overall combination.}
The convex-combination weights in the main-text formula
$D(e)=w_s D_s(e)+w_l D_l(e)+w_x D_x(e)$ are
$(w_s,w_l,w_x)=(0.55,\,0.35,\,0.10)$.  The A/B/C/D grade
thresholds applied to $D(e)$ are $(0.65,\,0.50,\,0.35)$.  All
constants in this appendix are coarse reporting choices for the
diversity histogram in Figure~\ref{fig:audit_summary}; they are not
learned and are not used by the RL trainer.  The audit additionally
writes every raw per-signal value, so flagged environments can be
inspected without relying on the aggregate grade alone.

\section{Example Environment Implementation}
\label{app:example_env}

For reference, Listing~\ref{lst:clock_angle} shows the full source
of one \tron{} environment from the Math bucket. The level ladder
(\texttt{\_level\_config}) selects question types as the difficulty
level $\ell$ increases; the generator (\texttt{\_generate\_problem})
samples a clock state, renders it with matplotlib, and returns a
\texttt{(question, answer, image)} triple. Numerical verification
with absolute tolerance $0.001$ is inherited from the base class
\texttt{StandaloneVisualEnv}.

\begin{lstlisting}[caption={Full source of \texttt{clock\_angle\_qa.py}.},label={lst:clock_angle}]
"""
Clock angle QA -- analog clock showing a time.
Questions: angle between hands, time shown, angle after N minutes, overlap count.
"""
import math
from typing import Dict, Optional, Tuple
import matplotlib; matplotlib.use("Agg")
import matplotlib.pyplot as plt
import numpy as np
from PIL import Image
from .standalone_base import StandaloneVisualEnv

class ClockAngleQA(StandaloneVisualEnv):
    ALLOW_ROTATION = False  # orientation-sensitive: disable rotation augmentation
    BENCHMARK_NUM_TOLERANCE_ABS = 0.001
    ENV_NAME = "clock_angle"

    def _hand_angle(self, h, m):
        """Return (hour_deg, minute_deg) measured clockwise from 12."""
        min_deg = 6 * m
        hour_deg = 30 * (h % 12) + 0.5 * m
        return hour_deg, min_deg

    def _angle_between(self, h, m):
        hd, md = self._hand_angle(h, m)
        diff = abs(hd - md)
        return min(diff, 360 - diff)

    def _level_config(self, level: int) -> Dict:
        level = max(0, min(level, 9))
        if level <= 2:
            return {"qtypes": ["read_time", "minute_hand_angle"]}
        if level <= 5:
            return {"qtypes": ["read_time", "angle_between",
                               "minute_hand_angle"]}
        if level <= 7:
            return {"qtypes": ["angle_between", "angle_after_n"]}
        return {"qtypes": ["angle_after_n", "overlap_count"]}

    def _generate_problem(self, seed: int, parameter: Dict) -> Optional[Tuple[str, str, Image.Image]]:
        rng = self._rng
        level = int(parameter.get("level", 0))
        cfg = self._level_config(level)
        style = self._random_style()
        qtype = parameter.get("question_type", rng.choice(cfg["qtypes"]))

        h = rng.randint(1, 12)
        m = rng.choice([0, 5, 10, 15, 20, 25, 30, 35, 40, 45, 50, 55])

        hd, md = self._hand_angle(h, m)

        sc = style["figsize_scale"]
        fig, ax = plt.subplots(figsize=(5 * sc, 5 * sc))
        fig.patch.set_facecolor(style["bg_color"])
        ax.set_facecolor(style["bg_color"])

        clock_face = plt.Circle((0, 0), 1.05, fc="white", ec=style["geo_line_color"],
                                linewidth=style["line_width"] + 1)
        ax.add_patch(clock_face)

        # Hour markers
        for i in range(1, 13):
            ang = math.radians(90 - 30 * i)
            ax.text(0.85 * math.cos(ang), 0.85 * math.sin(ang), str(i),
                    ha="center", va="center", fontsize=style["font_size_base"],
                    fontweight="bold", fontfamily=style["font_family"])
            ax.plot([0.95 * math.cos(ang), 1.0 * math.cos(ang)],
                    [0.95 * math.sin(ang), 1.0 * math.sin(ang)],
                    color=style["geo_line_color"], linewidth=1.5)

        # Hour hand
        h_ang = math.radians(90 - hd)
        ax.plot([0, 0.5 * math.cos(h_ang)], [0, 0.5 * math.sin(h_ang)],
                color=style["palette"][0], linewidth=style["line_width"] + 2,
                solid_capstyle="round")

        # Minute hand
        m_ang = math.radians(90 - md)
        ax.plot([0, 0.75 * math.cos(m_ang)], [0, 0.75 * math.sin(m_ang)],
                color=style["palette"][1], linewidth=style["line_width"] + 0.5,
                solid_capstyle="round")

        ax.plot(0, 0, "o", color=style["geo_line_color"], markersize=5, zorder=5)
        ax.set_xlim(-1.3, 1.3); ax.set_ylim(-1.3, 1.3)
        ax.set_aspect("equal"); ax.axis("off")
        ax.set_title("Analog Clock", fontsize=style["font_size_base"] + 2,
                     fontweight="bold")
        img = self.fig_to_pil(fig, dpi=style["dpi"])

        time_str = f"{h}:{m:02d}"
        angle = round(self._angle_between(h, m), 1)

        if qtype == "angle_between":
            q = ("For the time shown on the clock, what is the angle (in "
                 "degrees) between the hour and minute hands?")
            return q, str(angle), img
        elif qtype == "read_time":
            q = "What time is shown on the clock? Answer in H:MM format."
            return q, time_str, img
        elif qtype == "angle_after_n":
            dm = rng.choice([15, 30, 45, 60])
            new_m = (m + dm) % 60
            new_h = h + (m + dm) // 60
            new_angle = round(self._angle_between(new_h, new_m), 1)
            q = (f"Starting from the time shown on the clock, what will the "
                 f"angle between the hands be after {dm} minutes? Round to "
                 f"1 decimal.")
            return q, str(new_angle), img
        elif qtype == "overlap_count":
            n_hours = rng.choice([6, 12, 24])
            overlaps = round(n_hours * 11 / 12)
            if n_hours == 12:   overlaps = 11
            elif n_hours == 24: overlaps = 22
            elif n_hours == 6:  overlaps = 5
            q = f"How many times do the clock hands overlap in {n_hours} hours?"
            return q, str(overlaps), img
        elif qtype == "minute_hand_angle":
            q = ("For the time shown on the clock, how many degrees has the "
                 "minute hand moved from 12? Answer as a number.")
            return q, str(round(md, 1)), img
        return None
\end{lstlisting}

\end{document}